\documentclass[lettersize,journal]{IEEEtran}
\usepackage[caption=false,font=footnotesize]{subfig}
\usepackage{amsmath,amsfonts}
\usepackage{algorithmic}
\usepackage{algorithm}
\usepackage{array}
\usepackage{textcomp}
\usepackage{stfloats}
\usepackage{url}
\usepackage{verbatim}
\usepackage{graphicx}
\usepackage{cite}
\usepackage{orcidlink}
\usepackage{xcolor, color, soul, colortbl}
\usepackage{makecell}
\usepackage{nicematrix}
\usepackage{multirow}
\usepackage{float} 
\usepackage{booktabs}
\usepackage{textcase}
\usepackage{tabularx}
\usepackage{rotating}
\usepackage{siunitx}

\usepackage{etoolbox}
\makeatletter
\expandafter\patchcmd\csname\string\IEEEbiography\endcsname
  {\vskip \@IEEEBIOskipN plus 1fil minus 0\baselineskip}
  {\vskip 0.9\baselineskip}
  {}{}
\makeatother
 
\begin{document}

\title{PiLLVD: A Real-World Low-Light Video Dataset for Dynamic Scenes Captured with Off-the-Shelf Cameras}

 \author{Yaser Abu Awwad \orcidlink{0009-0008-0906-2415}, Paul L. Rosin \orcidlink{0000-0002-4965-3884}, Oktay Karakus \orcidlink{0000-0001-8009-9319}



\thanks{Yaser Abu Awwad, Paul L. Rosin, and Oktay Karakus are with the
School of Computer Science and Informatics, Cardiff University,
Cardiff CF24 4AG, U.K.
(Corresponding author: Yaser Abu Awwad;
e-mail: abuawwadyw@cardiff.ac.uk). \par The PiLLVD dataset is available on
Zenodo \cite{PiLLVD}.}}




\maketitle

\begin{abstract}
Low-light image and video enhancement has received increasing attention in recent years. However, most existing enhancement methods are developed and evaluated on still images, while progress in low-light video enhancement remains limited by the lack of real-world paired video datasets. Acquiring paired low-light and normal-light video sequences is challenging because it requires spatially and temporally aligned captures while avoiding motion blur, exposure mismatch, and correspondence errors in dynamic scenes. To address this gap, we introduce PiLLVD, the Pi-based Low-Light Video Dataset, a paired low-light video dataset captured using an accessible Raspberry Pi-based multi-camera system with off-the-shelf cameras. PiLLVD provides paired low-light and aligned normal-light reference sequences for dynamic indoor and outdoor scenes, including diverse human and object motion under varying illumination conditions. Unlike existing datasets that are often image-based, static, synthetic, or limited to simple motion patterns, PiLLVD focuses on motion-rich scenes where frame-to-frame stability and motion-related artefacts are important for evaluation. In addition to presenting the dataset and acquisition pipeline, we benchmark representative pre-trained low-light video enhancement methods on PiLLVD under a zero-shot setting. The results show that existing methods do not generalise uniformly across the proposed sequences, highlighting the challenges posed by the proposed dataset and its value for both benchmarking existing methods and supporting the development of future low-light video enhancement algorithms.
\end{abstract}


\section{Introduction}

Enhancing visual information captured under low-light conditions remains a persistent challenge across various fields, including anomaly detection in surveillance \cite{anomaly}, autonomous systems \cite{sid_2018}, underwater imaging \cite{underwater} and nighttime wildlife observation \cite{wildlife}.
Poor illumination typically results in increased noise, reduced contrast, and motion artefacts, all of which can degrade the quality and reliability of downstream processing tasks. Although a variety of image enhancement techniques ranging from traditional methods \cite{clahe_1994, retinex_1997} to learning-based approaches \cite{lightennet_2018, sid_2018, zero-dce_2020} have been proposed, they are predominantly developed and trained using still image datasets \cite{lol_2018, sice_2018}. These still-image datasets do not provide temporal information, making them unsuitable for evaluating video enhancement tasks where frame-to-frame coherence is critical.
As a result, models trained solely on still images often struggle to generalise to videos, as frame-independent enhancement can lead to inconsistent brightness, colour shifts, and flickering across consecutive frames.

\begin{figure*}[t]
  \centering
  \subfloat[DRV~\cite{DRV}\label{fig:drv}]{
    \includegraphics[width=0.2\textwidth]{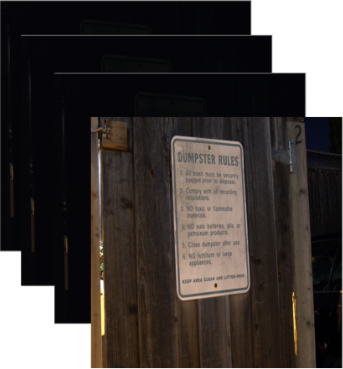}
  }\hfill
  \subfloat[SDSD~\cite{SDSD}\label{fig:sdsd}]{
    \includegraphics[width=0.25\textwidth]{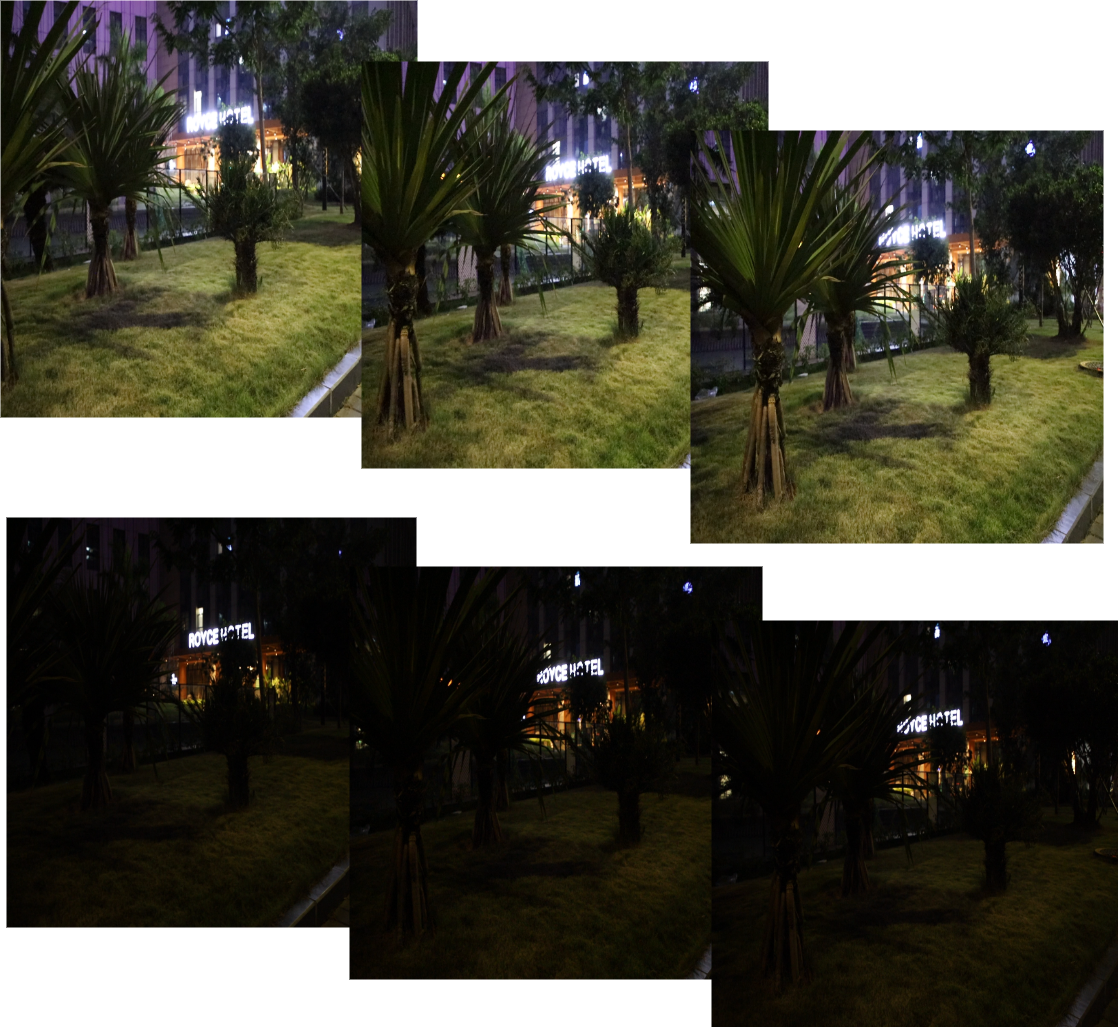}
  }\hfill
  \subfloat[Ours.\label{fig:ours}]{
    \includegraphics[width=0.4
    \textwidth]{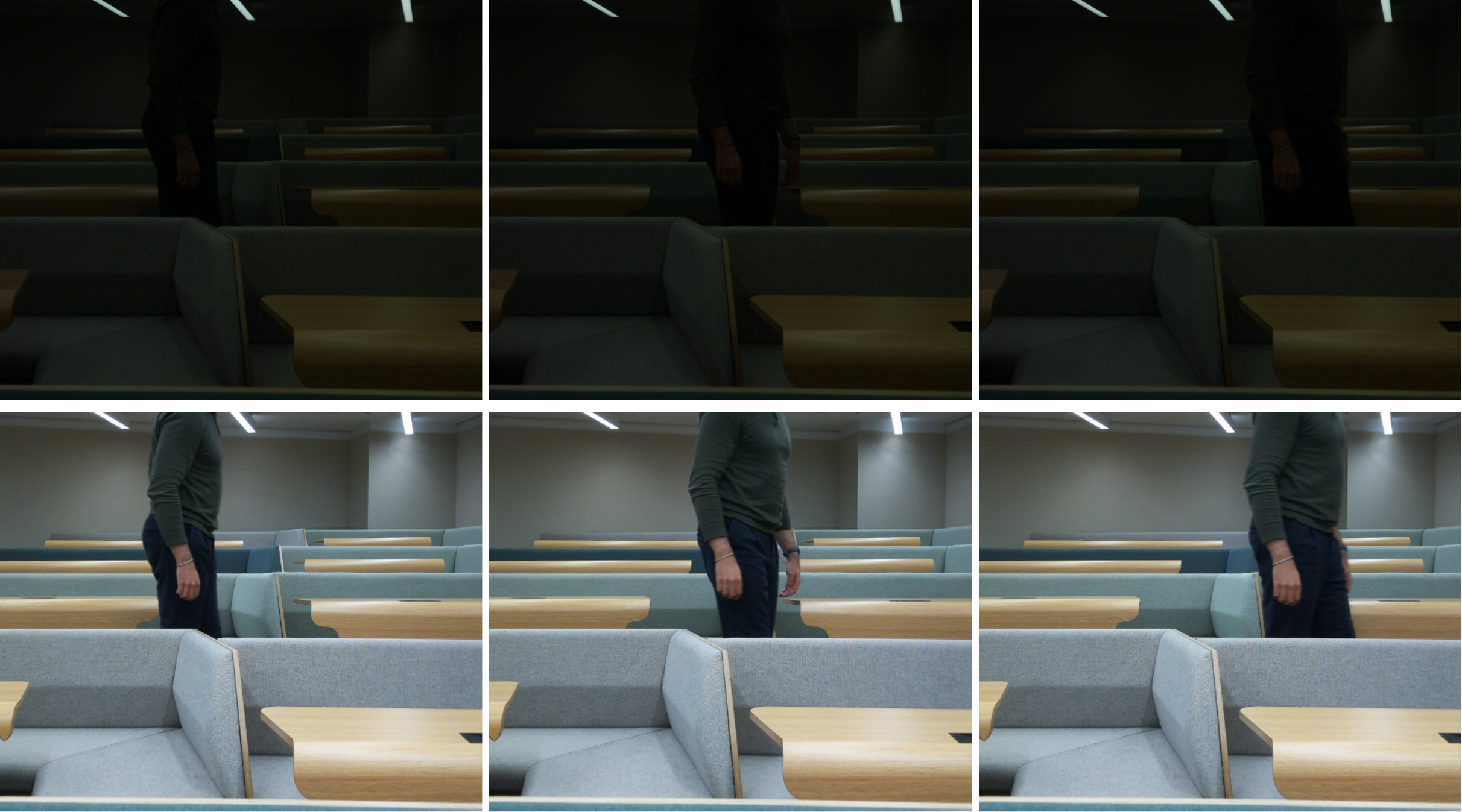}
  }
\caption{(a), (b) Sample frames from existing paired low-light video datasets; (c) sample frames from our dataset, shown from left to right as (\textit{frame0080, frame0090, and frame0100}), illustrating noticeable object motion across time.}
  \label{fig:data_comp}
\end{figure*}

Furthermore, only a limited number of video enhancement methods have been introduced in recent years. Some approaches \cite{StableLLVE, LVE-S2D_2022} attempt to adapt models trained on static images or synthetic video data \cite{VLN_2021, chhirolya_2022, LLVE-SEG_2023}. However, these strategies fail to capture the complexity of natural motion and temporal variation. Notably, some existing datasets that include realistic
object motion under low-light conditions are not publicly
accessible \cite{SMOID}, further limiting research progress
in this area. Consequently, existing video enhancement models, although designed for dynamic input, often do not reflect the challenges of real-world scenarios due to limitations in training data. The limitations of existing paired low-light video datasets and the dynamic nature of our proposed dataset are illustrated in Figure~\ref{fig:data_comp}. In particular, the sample frames from our dataset show noticeable object motion across successive frames, highlighting its suitability for both the development and benchmarking of low-light video enhancement methods under dynamic real-world conditions.

Collecting paired low-light video data also presents a significant system-level challenge that extends beyond algorithmic development. Achieving precise temporal synchronisation between cameras typically requires specialised hardware such as synchronised shutters, calibrated optical paths, or high-speed triggering modules. These setups are costly, difficult to reproduce, and often dependent on laboratory-grade equipment, limiting widespread accessibility. As a result, most existing studies rely on bespoke setups, making it difficult for the community to replicate data or extend datasets with more diverse scenes. A reproducible framework that uses widely available hardware can facilitate consistent benchmarking, collaborative dataset expansion, and fair comparison across enhancement methods. Such an approach contributes not only new data but also a practical foundation for long-term community-driven development.

In this paper, we tackle the limited availability of paired dynamic low-light video data by introducing PiLLVD, a real-world paired low-light video dataset captured using a scalable Raspberry Pi-based acquisition framework with off-the-shelf cameras.
Our goal is to bridge the gap between static image datasets and real-world video enhancement needs by constructing a dataset that captures temporally aligned sequences under varying low-light conditions.
The proposed system enables the acquisition of synchronised, motion-rich low-light and normal-light video pairs without relying on expensive laboratory-grade camera systems, while maintaining a low-cost setup. 
This makes it feasible to train and evaluate enhancement models that are better suited for dynamic, real-world environments.

Beyond serving as a benchmark for low-light video enhancement, PiLLVD can also support the development of future enhancement methods. The dataset spans varied motion patterns, illumination levels, and scene types across both indoor and outdoor environments. Its temporally aligned paired sequences allow the evaluation of frame-wise reconstruction quality, temporal stability, and robustness under realistic motion. The dataset may also support future downstream studies, including detection, tracking, and activity analysis in low-light conditions, and offers a realistic testbed for methods that jointly optimise enhancement and high-level perception. In addition, the diversity of scenes and motions makes PiLLVD suitable for investigating generalisation across domains and assessing the reliability of video models under challenging illumination.

The main contributions of this work are summarised as follows:

\begin{itemize}

    \item We develop a cost-effective Raspberry Pi-based acquisition system using off-the-shelf cameras and a beam splitter to capture synchronised paired low-light and normal-light video sequences.
    
    \item We introduce PiLLVD, a real-world paired low-light video dataset containing dynamic indoor and outdoor scenes with diverse human and object motion, varied illumination levels, and aligned normal-light reference frames.
    
    \item We provide a post-capture alignment and evaluation pipeline to improve spatial correspondence between paired sequences and assess the reliability of the aligned references.
    
    \item We benchmark representative pre-trained low-light video enhancement methods on PiLLVD under a zero-shot setting, highlighting the limitations of existing models when applied to unseen real-world dynamic low-light videos.
    
\end{itemize}

The rest of this paper is structured as follows: Section \ref{rw} reviews related work; Section \ref{dataset} describes our system design and data collection method; Section \ref{exp} presents experimental results and benchmarking; Section \ref{Dis} discusses the findings and limitations; and Section \ref{con} concludes the paper.

\section{Related Work}
\label{rw}

This section reviews existing low-light datasets for images and videos, followed by an overview of low-light video enhancement methods.

\subsection{Low-Light Datasets}

\subsubsection{Image datasets} One of the earliest efforts to support low-light image enhancement was the LOw-Light (LOL) dataset \cite{lol_2018}, which provides paired low-light and well-lit images comprising both real and synthetic samples. The real images were captured under controlled illumination conditions, while the synthetic counterparts were generated by transferring low-light characteristics from datasets such as MEF \cite{MEF}, NPE \cite{NPE}, LIME \cite{LIME}, and DICM \cite{DICM} to well-lit images from the RAISE dataset \cite{RAISE} using Adobe Lightroom. This combination introduces diverse lighting conditions essential for training supervised models.
In parallel, the SID dataset \cite{sid_2018} presented a collection of short-exposure and long-exposure image pairs captured in extremely low-light environments. Each long-exposure image is paired with multiple short-exposure counterparts taken under varying conditions. Operating at the raw image level, this dataset has become a benchmark for evaluating low-light image restoration techniques.
The SICE dataset \cite{sice_2018} extends this approach by capturing multiple exposure levels for each scene in both indoor and outdoor environments. High-quality reference images were generated using fusion-based and HDR techniques, followed by manual selection to ensure perceptual quality. This dataset is particularly well-suited for assessing contrast and exposure enhancement algorithms.
In a related effort, the MIT-Adobe FiveK dataset \cite{MIT_Adobe} comprises photographs retouched by professional photographers, each offering several enhanced versions exhibiting different visual styles. Although not specifically designed for low-light conditions, its wide range of degradations and adjustments makes it valuable for general image enhancement tasks.
More recently, the VE-LOL dataset \cite{VE-LOL} was introduced to support both low-level and high-level vision tasks. VE-LOL-L contains real and synthetic low-light images for enhancement and denoising, while VE-LOL-H includes unpaired facial and object images captured under various lighting conditions for applications such as detection and recognition. Additionally, unpaired datasets such as ExDark \cite{ExDark}, MEF, and NPE provide real-world low-light images captured under diverse conditions, including nighttime and twilight. These datasets are commonly used in unsupervised and zero-shot enhancement approaches, particularly where ground-truth references are unavailable. 

Although these datasets have significantly advanced low-light image enhancement, they are not designed for video enhancement. They do not provide temporally continuous paired frames and therefore cannot support the evaluation of temporal stability, motion-related artefacts, or frame-to-frame consistency. This limits their suitability for developing low-light video enhancement methods intended for dynamic real-world scenes.

\subsubsection{Video datasets} Building upon image datasets, low-light video datasets play a crucial role in advancing video enhancement research. Unlike still images, video data introduces additional challenges, such as maintaining temporal consistency and mitigating motion blur and noise across consecutive frames. These factors make the acquisition of high-quality low-light video datasets particularly complex.
One of the first attempts to construct a paired dataset for this purpose was the Dark Raw Video (DRV) dataset \cite{DRV}, which provides both low-light and normal-light video pairs captured under controlled lighting conditions. The dataset features real-world static scenes, such as stationary objects with no camera or object motion, and offers a foundation for learning raw-to-sRGB mappings in low-light environments. However, the absence of scene dynamics limits its suitability for training or evaluating models intended for real-world, motion-rich video scenarios.
The SMOID dataset \cite{SMOID} extends this by focusing on dynamic scenes involving object and pedestrian motion. To simulate low-light conditions, the authors used identical cameras with different optical settings, including the application of a neutral density (ND) filter and varying analogue gain values. This setup enabled side-by-side capture of the same scene under both lighting conditions, facilitating alignment between frames and ensuring consistent content across modalities and supporting tasks such as motion-aware enhancement and object tracking. However, the dataset has not been made publicly available, which limits its accessibility and hinders the development and benchmarking of low-light video enhancement methods. As a result, the full range of motion patterns and scene variation cannot be independently examined.

SIDGAN~\cite{SIDGAN} synthesises long/short-exposure pairs from web videos via a dual-CycleGAN that inserts a long-exposure “bridge” domain. These pairs, combined with limited real data, are used to train a RAW-to-RGB low-light video model with improved colour fidelity and temporal consistency. However, synthetic-only training underperforms, and the assessment leans heavily on static scenes.
In a similar direction to \cite{SMOID}, the SDSD dataset \cite{SDSD} captures both indoor and outdoor dynamic scenes using a dual-camera setup mounted on an electric rail to maintain consistent movement. Low-light conditions were introduced using ND filters in combination with controlled ambient lighting and different analogue gain settings. This setup allows for realistic simulation of camera motion and scene dynamics under reduced illumination, making the dataset well-suited for temporal modelling in enhancement algorithms. However, although the scenes are dynamic, the camera movement is constrained to a fixed linear path, which does not fully reflect the variability and unpredictability of real-world handheld or multi-camera scenarios. In addition, the motion mainly arises from controlled camera translation rather than diverse human or object activity, limiting the variety of dynamic behaviours represented in the dataset.
The authors in \cite{R-DID} introduced two paired low-light video datasets, DID and R-DID, captured using different rigs to prioritise either spatial alignment (DID) or temporal realism (R-DID). DID was captured frame-by-frame at the same location, where a central controller adjusted the ISO to obtain a low-light frame and its corresponding normal-light counterpart. The controller verified image quality, re-shot if necessary, and slightly rotated the gimbal (by less than 1$^{\circ}$) to introduce subtle motion while maintaining scene continuity. In essence, the low-light condition was achieved by reducing exposure through ISO relative to the normal-light setting.
Conversely, R-DID involved continuous dual recording using two side-by-side smartphones. The controller simultaneously started and stopped both devices, configuring one to record in low-light and the other in normal light. Afterwards, each frame pair was spatially aligned to construct the paired video sequences. Although the data are dynamic, low light is simulated mainly by adjusting ISO, which may not fully reflect naturally occurring low-light acquisition conditions. The DID dataset captures static scenes, whereas R-DID addresses dynamic content but retains notable limitations. Owing to the absence of a precise capture system and the use of side-by-side cameras, substantial misalignment can occur between paired frames, forcing the alignment algorithm to apply considerable warping during post-processing. Moreover, the resolution of the dynamic dataset is relatively low. Since the datasets are not publicly accessible, the full diversity of motion patterns, scene content, and dynamic activities cannot be independently assessed.

At the time of writing, BVI-RLV \cite{BVI} represents one of the most recent datasets for low-light video enhancement that provides paired low-light and normal-light videos. Unlike simultaneous dual-camera capture, it employs a programmable motorised dolly to generate repeatable motion and records normal-light videos as pseudo ground truth. These videos are then aligned to the low-light sequences through a frame registration process to construct paired data. 
Although this design improves spatial and temporal correspondence between pairs, the captured motion is predefined through a motorised dolly with fixed movement settings, and the diversity of moving objects, scene content, and human activities remains limited. As a result, it may not fully reflect the variability of real-world dynamic scenes.
Furthermore, the dataset is acquired under fixed low-light levels, potentially limiting the diversity of illumination conditions. 

Overall, existing low-light datasets remain limited for dynamic video enhancement. Image datasets lack temporal structure, while many video datasets are static, synthetic, constrained by predefined motion, or not publicly accessible. These limitations reduce their ability to support the evaluation of frame-to-frame stability under natural motion, motion-related artefacts, and generalisation under diverse illumination conditions. PiLLVD is designed to address this gap by providing paired real-world low-light and normal-light video sequences containing diverse dynamic indoor and outdoor scenes. Table~\ref{tab:datasets} summarises the main differences between PiLLVD and related real dynamic paired low-light video datasets. Although PiLLVD is smaller in frame count, it focuses on uncontrolled human and object motion across both indoor and outdoor environments, with higher motion diversity and variable high-resolution frames. This comparison highlights the complementary role of PiLLVD and motivates its use for both benchmarking existing methods and developing future low-light video enhancement models under dynamic real-world conditions.

\begin{table*}[t]
\renewcommand{\arraystretch}{1.4}
\centering
\caption{A detailed comparison of our dataset with existing real dynamic paired datasets.}
\label{tab:datasets}
\resizebox{\textwidth}{!}{%
\begin{tabular}{ccccccccc}
\toprule
Dataset &  Motion source & Motion pattern & Motion diversity &  Camera & Environment & Resolution & No. of frames & Released \\
\midrule

SMOID & Human \& Object & Uncontrolled & 
Unknown & FLR GS3-U3-23S6C  & Indoor & 1800 $\times$ 1000  & 35,800 & No  \\[3pt]

SDSD  &  Camera & Controlled & Low & Canon 6D Mark II & Outdoor &   \makecell{512 $\times$ 960 \\ 1920 $\times$ 1080} & 37,800 & Yes   \\[9pt]


R-DID  & Human \& Object & Uncontrolled & Unknown & HUAWEI P40 Pro & Indoor \& Outdoor &  1280 $\times$ 720    & 41,038 & No    \\[9pt]

BVI-RLV & Camera \& Object & Controlled & Low &  Sony Alpha 7SII & Indoor & 1920 $\times$ 1080 & 31,800 & Yes  \\
\midrule
\textbf{PiLLVD} &  Human \& Object & Uncontrolled & High & Arducam IMX519 & Indoor \& Outdoor &  Variable & 4,531 & Yes  \\
\bottomrule
\end{tabular}}
\end{table*}

\subsection{Low-light Video Methods}
Low-light video enhancement (LLVE) aims to improve visibility, contrast, colour fidelity, and perceptual quality in videos captured under insufficient illumination. Unlike single-image enhancement, LLVE must also preserve temporal consistency across consecutive frames. Enhancing each frame independently may improve brightness, but it can introduce flickering, unstable colours, temporal noise, and inconsistent details, especially in dynamic scenes with object motion, camera motion, and illumination variation.
Existing enhancement methods can be broadly grouped into supervised, unsupervised, semi-supervised, zero-reference, and video-aware approaches. Supervised methods learn from paired low-light and normally exposed data, with examples including LLNet \cite{LLNet}, MBLLEN/MBLLVEN \cite{mbllen_2018}, and Retinex-inspired models such as KinD/KinD++ \cite{kind++_2021}. These methods benefit from direct reference supervision, but their performance depends heavily on the availability and diversity of accurately aligned paired data. This is particularly challenging for real-world videos, where low-light and normal-light pairs must remain spatially and temporally consistent despite motion.

To reduce the dependence on paired data, unsupervised, semi-supervised, and zero-reference methods have also been proposed. EnlightenGAN \cite{EnlightenGAN} learns from unpaired low-light and normal-light images, while semi-supervised methods combine limited paired supervision with unpaired data \cite{semi_2020}. Zero-reference methods such as Zero-DCE \cite{zero-dce_2020} estimate enhancement functions using non-reference constraints, and SGZ extends zero-shot enhancement to low-light image/video enhancement using semantic guidance \cite{SGZSL_2022}. Although these approaches improve applicability when reference data are unavailable, they are often image-centred and may produce over-enhancement, colour shifts, or temporal inconsistency when applied to videos in practical dynamic video scenarios.
Video-aware LLVE methods explicitly exploit temporal information by using neighbouring frames, 3D convolutions, recurrent modules, optical-flow alignment, transformers, or temporal consistency losses. MBLLVEN uses 3D convolutions for low-light videos \cite{mbllen_2018}, while other recent methods focus on improving temporal stability or jointly addressing related degradations such as denoising and deblurring \cite{LEDNet,VJT}. These approaches better reflect practical low-light video restoration, where poor illumination is often accompanied by sensor noise, blur, and motion-related artefacts.

Overall, LLVE methods have progressed from hand-crafted enhancement to learning-based and video-aware restoration. However, many models are still adapted from image enhancement or trained on datasets with limited motion diversity. This highlights the need for paired low-light video datasets containing realistic motion, illumination variation, noise, blur, and temporal changes. Such datasets are essential not only for benchmarking existing methods, but also for supporting the development of future models that are visually effective, temporally stable, and robust in complex real-world low-light videos.

\begin{table}[!b]
   \caption{Details of system components.}
    \label{tab:sys}
   \centering
   \resizebox{\columnwidth}{!}{%
    \begin{tabular}{ccc}
    \toprule
    Component & Description & Quantity \\
    \midrule
    Raspberry Pi &  Model 4B/4GB  &  3  \\ \\
    Beam Splitter & \makecell{\textbf{Dimension:} $50 \text{mm}\times 50 \text{mm}$ \\ \textbf{Thickness:} $3.00\text{mm}$ \\ \textbf{Shape:} Plate \\ \textbf{Angle of incident:} $45\text{\textdegree}$ \\
    \textbf{Reflection/Transmission}: $70/30$ \\ 
    \textbf{Wavelength:} $400-700\text{nm}$  \\
    \makecell{\textbf{Anti-reflection coating:} Yes}} & 1 \\ \\
    Camera & \makecell{Arducam Sony-IMX519} & 2 \\
    \bottomrule
  \end{tabular}}
\end{table}



\begin{figure*}[t]
    \centering
    \subfloat[]{
        \includegraphics[width=\columnwidth]{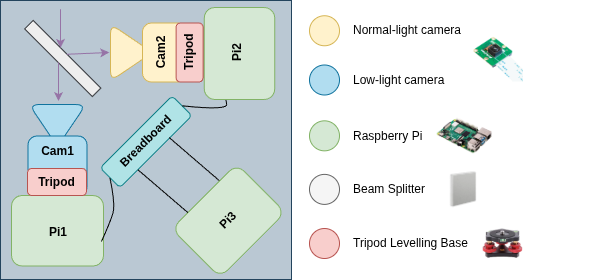}%
        \label{fig:sys_a}
    }
    \subfloat[]{
        \includegraphics[width=0.6\columnwidth]{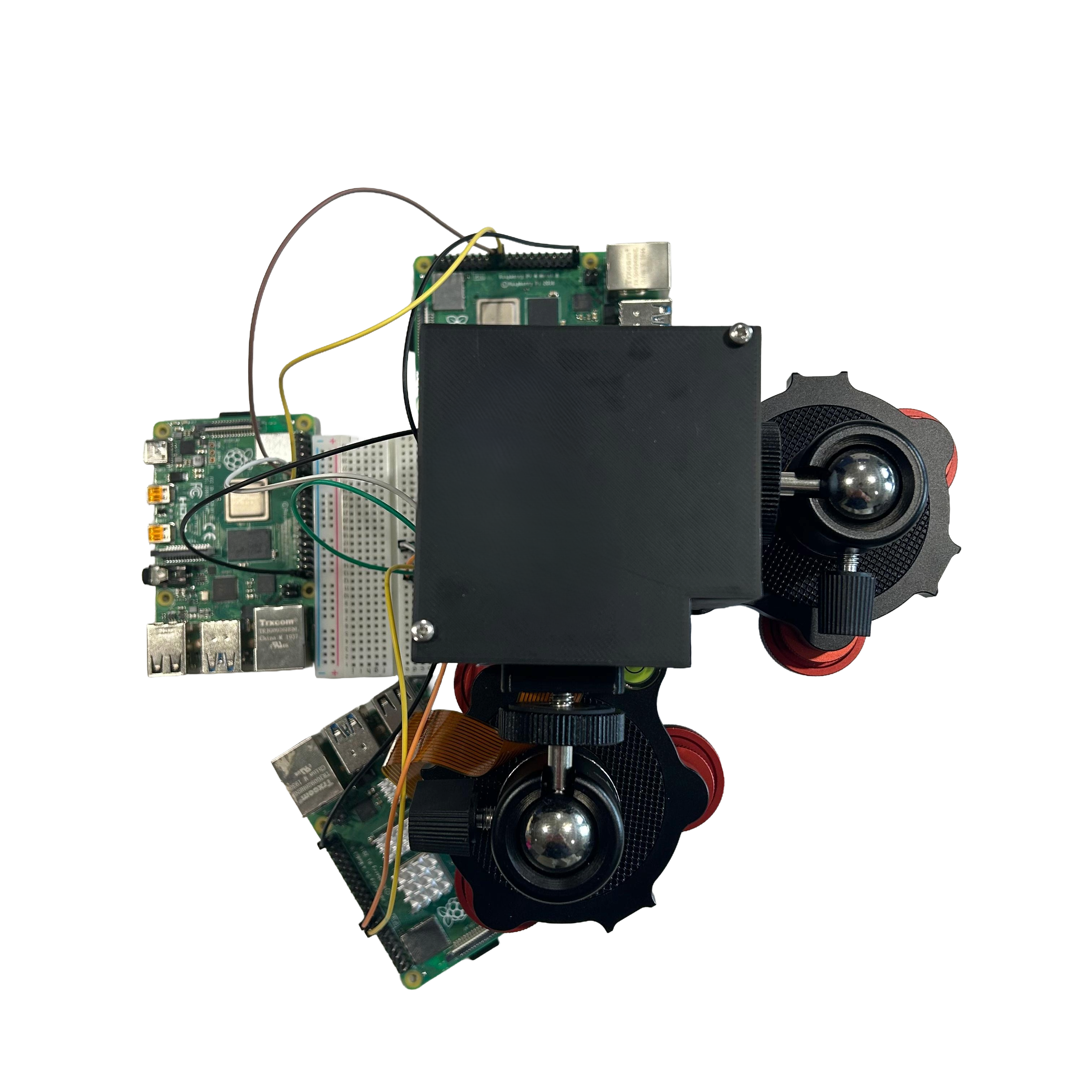}%
        \label{fig:sys_b}
    }
    \caption{The proposed camera setup for capturing paired low-light and normal-light video frames. (a) Schematic diagram showing the optical path and components used. (b) The actual system.}
    \label{fig:sys}
\end{figure*}

\section[The PiLLVD Dataset]{The P{\normalfont i}LLVD Dataset}
\label{dataset}

\subsection{System Setup}

To capture paired low-light and normal-light video sequences for dynamic scenes, we built a synchronised dual-camera system with a beam splitter, as shown in Fig.~\ref{fig:sys}. In contrast to high-cost and highly specialised systems used in previous studies, our setup is designed to be low-cost, scalable, and easy to reproduce using off-the-shelf components.

\subsubsection{Hardware} Our proposed system is composed of several components, including cameras, a beam splitter, and single-board computers (RPis). These components are enclosed in a custom-designed 3D-printed mount to ensure fixed positioning, alignment stability, and ease of deployment. 
The cameras are identical low-cost modules; one captures the low-light view, while the other captures the corresponding normal-light reference view. The beam splitter divides the incident light into two paths: 30\% of the light is transmitted to the low-light camera, while the reflected 70\% path is directed to the normal-light reference camera. Moreover, the resource-constrained devices serve as the core of the system, performing key functions including image capture, processing, and task management.
The Pi3 functions as the master device, while Pi1 and Pi2 operate as slaves. The master serves as a Network Time Protocol (NTP) server, distributing timestamps to the slave devices for time synchronisation. Additionally, it is responsible for sending pulse triggers to initiate and manage the image capturing process, thereby ensuring precise temporal alignment across devices. This triggering mechanism is designed to minimise the temporal offset between the two cameras by sending the capture signal simultaneously to both devices. The recorded capture timestamps were stored in text files and showed that the temporal offset between paired frames was below one millisecond. Data were captured frame by frame, with the camera operated in video mode to maintain continuous acquisition and stable capture settings across frames. A single trigger pulse is sent to both Pi1 and Pi2 simultaneously to initiate capture at the same time. All Raspberry Pis are interconnected through their GPIO pins and powered from 5V adapters, ensuring a common signalling interface and consistent power delivery. Table~\ref{tab:sys} provides detailed descriptions of the components used in our system.

\subsubsection{Pre-capture} Before starting the capturing process, several preparatory steps were carried out. First, each camera was mounted on a tripod levelling-base to allow for rotation and translation adjustments. We followed a strategy of displaying horizontal and vertical cursors on the live video stream and performing an initial alignment of each camera with a calibration target. This ensured that both cameras were directed at the same point in the scene while maintaining a small and consistent alignment error throughout the capturing process. It is important to note that manual alignment was performed through the beam splitter. We assume that refraction effects remain consistent during acquisition, particularly along the transmitted path.
Autofocus was used during setup to maximise sharpness, after which the focus was fixed during capture. For both devices, the resolution was set to (3840 $\times$ 2160), while ISO, analogue gain, exposure time, and colour gains were held constant to ensure cross-device comparability and consistent capture conditions.

\begin{table}[b]
\renewcommand{\arraystretch}{1.4}
\centering
\caption{Quantitative comparison of alignment performance before and after
alignment, averaged across all videos in the dataset.}
\label{tab:align_metrics}
\resizebox{\columnwidth}{!}{%
\begin{tabular}{ccc}
\hline
Metric & Before & After\\
\hline
Mean Matched Feature Displacement (px) $\downarrow$ & 18.790  & 0.560  \\
Mean Normalised Feature Displacement (\%) $\downarrow$ & \phantom{1}0.517 & 0.017 \\
Mean Reprojection Error (px) $\downarrow$ & \phantom{1}0.860  & 0.460   \\
Mean Normalised Reprojection Error (\%) $\downarrow$ & \phantom{1}0.024 & 0.014 \\
Mean Inlier Ratio $\uparrow$ & \phantom{1}0.770  & 0.931  \\
\hline
\end{tabular}%
}
\end{table}

\subsubsection{Post-capture} 
\begin{figure*}
    \centering
    \includegraphics[width=1\textwidth]{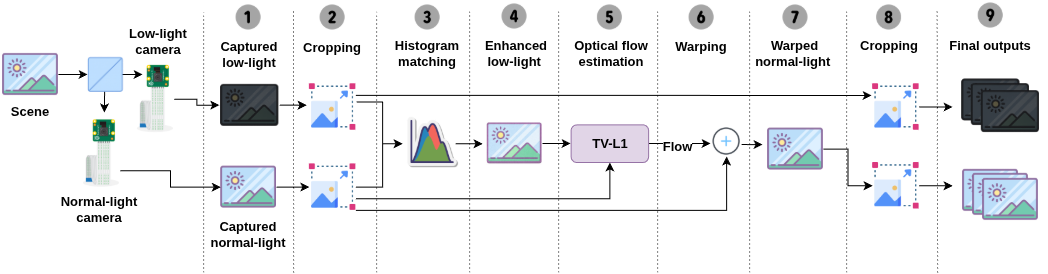}
    \caption{Overview of the proposed acquisition and post-processing pipeline for constructing paired low-light and normal-light data.}
    \label{fig:pp}
\end{figure*}

After capture, both low-light and normal-light frames were cropped to remove boundary artefacts caused by the 3D-printed enclosure.
Even though an initial alignment was performed during the pre-capture stage, geometric misalignment may remain due to small temporal offsets, physical mounting tolerances, viewpoint differences, lens distortion, and refraction through the beam splitter.
An additional alignment step is therefore required. However, aligning the paired frames is challenging due to the low-light conditions and noise in the dark frames, often resulting in imperfect alignment and the introduction of artefacts.
Thus, for the purpose of optical-flow estimation, we enhance the low-light frames $ll$ by matching their intensity distribution to the corresponding normal-light reference frames $nl$ using histogram matching \cite{hm}, producing $ll\_enhanced$.
It is worth mentioning that histogram matching is an intensity-only transformation, meaning the spatial structure of the image remains unchanged.
Afterwards, we applied the method proposed by \cite{of}  to estimate the optical flow between the normal-light ($nl$) and enhanced low-light $ll\_enhanced$ frames. The resulting flow was used to align the normal-light frames to the low-light ones, producing $nl\_warped$. This design preserves the low-light frames as originally captured and avoids introducing interpolation artefacts into the observations used as inputs to the enhancement algorithms. Finally, the paired frames were cropped to remove artefacts introduced during the warping stage. Figure~\ref{fig:pp} presents an overview of the proposed system and its processing pipeline.

\subsubsection{Alignment accuracy}

Alignment accuracy was evaluated using feature-based
geometric metrics by comparing matched feature
correspondences before and after alignment.
The evaluation used mean matched feature displacement
\cite{zitova_2003}, mean reprojection error
\cite{hartley_2004}, and mean inlier ratio
\cite{fischler_1981}. These metrics are commonly used in feature-based image
registration and geometric model estimation.
They measure both the residual spatial offset between
corresponding points and the consistency of the estimated
transformation.
Table~\ref{tab:align_metrics} summarises the quantitative
comparison before and after applying the proposed alignment
procedure, averaged across all videos in the dataset.
The mean matched feature displacement measures the average
pixel distance between corresponding matched features.
The mean reprojection error measures the residual error
between the observed feature locations and their positions
predicted by the estimated geometric transformation.
Since the final frame resolutions vary across sequences,
both errors are also reported in normalised form by dividing
the pixel error by the image diagonal. This allows fairer
comparison across videos with different spatial resolutions.
The mean inlier ratio indicates the proportion of feature
matches that are consistent with the estimated transformation.

\begin{figure}[b]
  \includegraphics[width=\columnwidth]{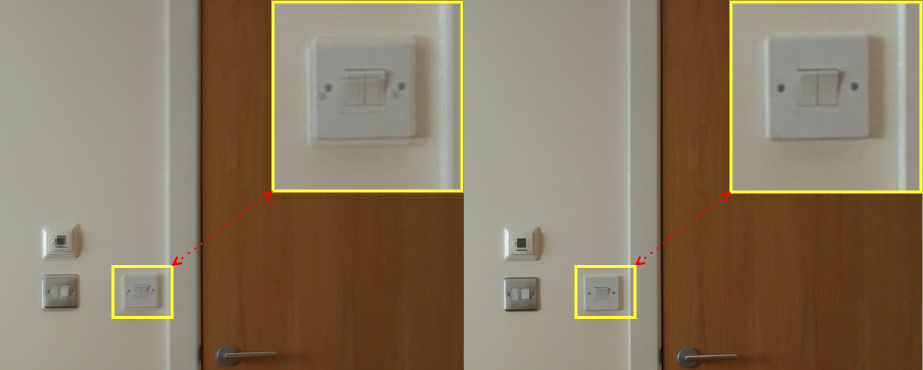}
  \caption{Comparison of normal-light and low-light frame fusion results before and after alignment. The marked regions highlight redundant scene content caused by misalignment in the pre-alignment fused image, which is reduced after alignment.}
  \label{v_align}
\end{figure}

In addition to the quantitative evaluation,
Fig.~\ref{v_align} provides a visual comparison of the fusion
results before and after alignment. 
The highlighted regions show that duplicated scene content caused by pre-alignment misregistration is reduced after applying the alignment procedure. This provides visual evidence of improved spatial
correspondence between the paired frames.

\subsection{Dataset}

Our dataset, PiLLVD, comprises 17 indoor and 7 outdoor paired video sequences, for a total of 24 paired sequences. Each sequence contains between 30 and 300 frames, with 4,531 frames in total. This count refers to each domain individually; the dataset therefore contains 4,531 low-light frames and 4,531 corresponding normal-light reference frames. All frames are stored in RGB format. Although PiLLVD is compact in scale, it is designed to maximise scene and motion diversity by including different objects, human activities, backgrounds, and indoor/outdoor illumination conditions. A comparison with related real dynamic paired low-light video datasets is provided in Table~\ref{tab:datasets}.
After sequence-specific cropping and alignment, the final frame resolutions range from \(1700 \times 1400\) to \(3110 \times 2160\) pixels, reflecting the use of different capture configurations across the recorded scenes. A sample from the dataset is shown in Figure~\ref{fig:ours}. 
PiLLVD is characterised by realistic scene dynamics, capturing natural motion from both humans and non-human objects under diverse low-light conditions. 
Human motion includes walking, hand movement, gestures, and partial body motion, while object motion includes handheld objects, moving toys, and screen content. All individuals appearing in the recorded sequences provided informed consent for their participation and for the use and public release of the captured video data for research purposes.

\begin{figure*}[t!]
    \centering
    \subfloat[Indoor]{
        \includegraphics[width=\columnwidth]{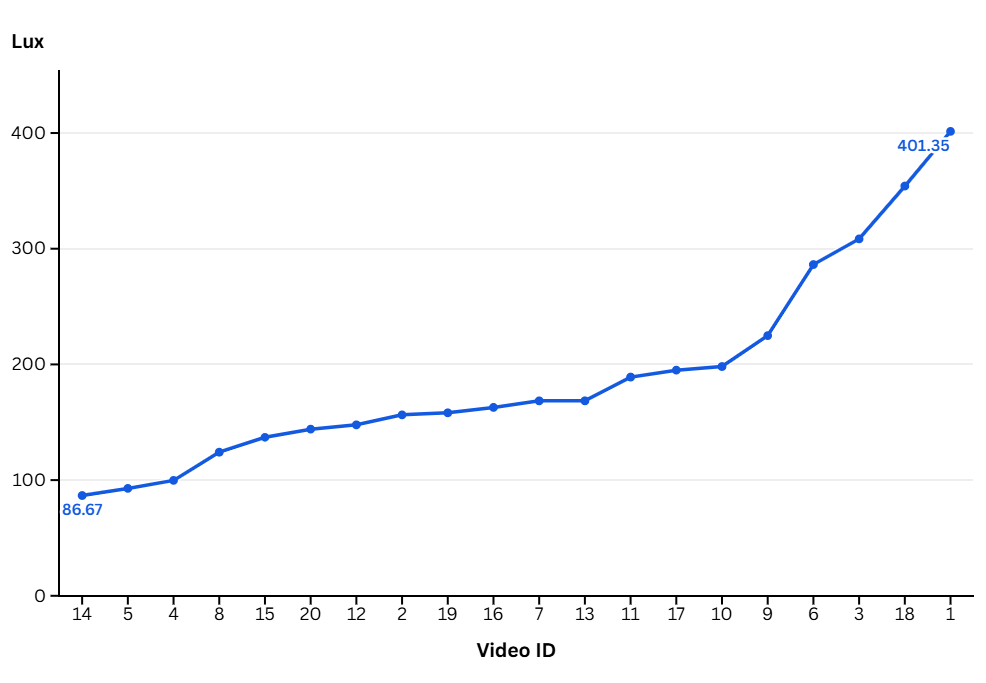}%
        \label{fig:lux_indoor}
    }
    \subfloat[Outdoor]{
        \includegraphics[width=\columnwidth]{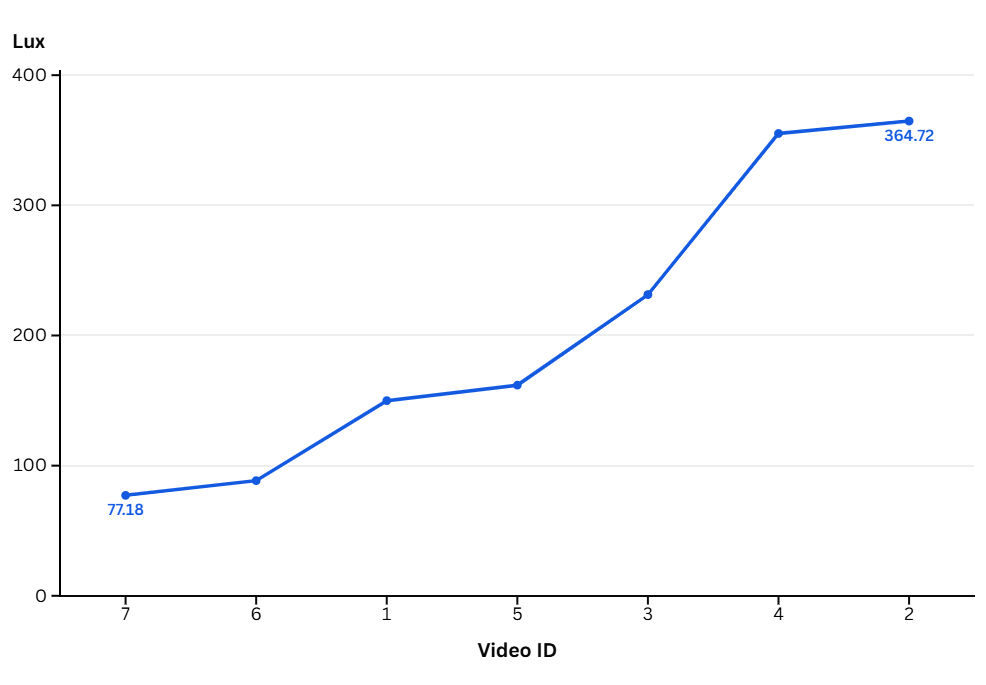}
        \label{fig:lux_outdoor}
    }
    \caption{Distribution of measured illuminance values for the low-light videos in the dataset.}
    \label{fig:lux_dist}
\end{figure*}

The illumination characteristics of PiLLVD are summarised in Figure~\ref{fig:lux_dist}, which shows the distribution of measured illuminance values for the low-light videos across indoor and outdoor sequences. These illuminance values were recorded during acquisition and stored in the accompanying metadata for each sequence. The measured values span approximately 77 to 401 lux, and the corresponding videos appear dark and exhibit typical low-light degradation, including reduced contrast, noise, and loss of detail. In terms of acquisition scenario, the cameras remain static while the recorded scenes contain dynamic motion. This setting is representative of practical fixed-camera applications, such as surveillance, monitoring, and indoor activity recording, where the camera viewpoint is fixed but people or objects move within the scene. Together, the illumination variation, dynamic content, and fixed-camera capture setup enable a robust evaluation of enhancement methods under realistic operating conditions. 
Future versions will expand PiLLVD with additional outdoor scenes, faster motion, and more challenging low-light conditions.

\section{Experiments and Results}
\label{exp}

\subsection{Experimental Objective}
\label{eo}

The objective of this section is to evaluate the generalisation capability of existing low-light video enhancement models on the proposed PiLLVD dataset. The purpose of this benchmark is not to introduce a new enhancement model, but to evaluate how existing pre-trained methods generalise to the dynamic paired videos in PiLLVD. Unlike training-based benchmarking, where models are adapted to the target dataset, all selected methods are assessed directly using their publicly available pre-trained weights, without any fine-tuning on PiLLVD. This zero-shot setting reflects a practical deployment scenario in which enhancement models trained on existing datasets are applied to unseen videos captured with different cameras, illumination conditions, scene content, and motion patterns.
Specifically, this evaluation aims to determine which pre-trained models can generalise most effectively to dynamic low-light video sequences. The low-light frames are used as inputs, while the corresponding spatially and temporally aligned normal-light frames are used as references for quantitative assessment. Since video enhancement involves both spatial restoration and temporal coherence, the benchmark evaluates reconstruction fidelity, perceptual quality, and stability across consecutive frames. In this way, PiLLVD provides a challenging testbed for analysing the robustness and practical applicability of existing enhancement methods under realistic low-light video conditions. More broadly, the benchmark demonstrates how PiLLVD exposes different failure modes of existing enhancement approaches when applied to dynamic paired low-light videos, supporting its role as both an evaluation resource and a basis for future method development.

\subsection{Benchmark Models}
\label{bm}

To evaluate the generalisation capability of existing enhancement methods on PiLLVD, we selected representative models that satisfy five criteria: (i) compatibility with RGB input frames, (ii) availability of public implementations and pre-trained weights, (iii) relevance to low-light video enhancement or video-compatible enhancement, (iv) applicability without requiring additional sensor modalities, such as RAW Bayer data or event-camera input, and (v) feasible inference under the available computational resources. These criteria allow all methods to be evaluated under a consistent zero-shot setting using the low-light frames from PiLLVD.

The selected methods are SDSDNet~\cite{SDSD}, DP3DF~\cite{DP3DF}, StableLLVE~\cite{StableLLVE}, SGLLIE/SGZ~\cite{SGZSL_2022}, and LLVE-STCD~\cite{LLVE_STCD}. These models cover different enhancement strategies, including supervised video enhancement, parametric spatio-temporal filtering, temporal consistency learning, semantic-guided zero-shot enhancement, and spatial-temporal decomposition. SDSDNet is included as a dataset-driven low-light video enhancement baseline, while DP3DF and LLVE-STCD represent spatio-temporal approaches that exploit information across frames. StableLLVE is selected for its focus on temporal consistency. SGLLIE/SGZ is included as a video-compatible zero-shot method, providing a reference-free comparison against supervised video-based models. Together, these methods enable PiLLVD to assess how existing pre-trained models handle unseen real-world motion, noise, illumination variation, and temporal instability.

It is important to note that the selected methods also differ in computational requirements and output resolution. In our experiments, several methods required GPU-based inference or operated on resized frames to make processing feasible. This highlights the need for efficient low-light video enhancement methods suitable for real-time or resource-constrained deployment.

\begin{table*}[t]
\centering
\caption{Quantitative comparison of different enhancement models on indoor PiLLVD sequences.}
\label{tab:QuanR_indoor}

\renewcommand{\arraystretch}{1.4}
\resizebox{\textwidth}{!}{%
\begin{tabular}{c|cccc|cccc|cccc|cccc}
\hline

\multirow{2}{*}{Model}
& \multicolumn{4}{c|}{PSNR $\uparrow$}
& \multicolumn{4}{c|}{SSIM $\uparrow$}
& \multicolumn{4}{c|}{LPIPS $\downarrow$}
& \multicolumn{4}{c}{mTWE $\downarrow$} \\
\cline{2-17}

& V4 & V5 & V14 & V17
& V4 & V5 & V14 & V17
& V4 & V5 & V14 & V17
& V4 & V5 & V14 & V17 \\
\hline

SDSDNet
& 22.839 & 21.909 & 18.843 & 19.232
& 0.584 & 0.642 & 0.604 & 0.601
& 0.212 & 0.180 & 0.198 & 0.212
& 0.028 & 0.026 & 0.023 & 0.022 \\

DP3DF
& 20.381 & 23.007 & 20.835 & 21.011
& 0.636 & 0.806 & 0.794 & 0.816
& 0.336 & 0.176 & 0.143 & 0.115
& 0.016 & \textbf{0.007} & \textbf{0.009} & \textbf{0.006} \\

LLVE-STCD
& 20.375 & \textbf{23.169} & \textbf{21.091} & \textbf{26.150}
& \textbf{0.750} & \textbf{0.862} & \textbf{0.860} & \textbf{0.910}
& 0.197 & \textbf{0.112} & \textbf{0.108} & \textbf{0.112}
& \textbf{0.013} & 0.008 & 0.010 & 0.007 \\

StableLLVE
& 18.660 & 22.892 & 20.420 & 20.575
& 0.694 & 0.798 & 0.790 & 0.785
& \textbf{0.182} & 0.195 & 0.216 & 0.203
& 0.030 & 0.016 & 0.019 & 0.018 \\

SGLLIE/SGZ
& \textbf{25.684} & 18.843 & 15.795 & 16.510
& 0.662 & 0.602 & 0.550 & 0.536
& 0.303 & 0.380 & 0.394 & 0.429
& 0.053 & 0.031 & 0.032 & 0.031 \\
\hline

\end{tabular}%
}
\end{table*}

\subsection{Evaluation Protocol}
\label{ep}

All selected models are evaluated directly on PiLLVD using their publicly available pre-trained weights, without retraining, fine-tuning, or domain adaptation. For each paired sequence, the low-light frames are used as the model inputs, while the corresponding spatially and temporally aligned normal-light frames are used as reference frames for full-reference evaluation. This protocol is intended to assess the zero-shot generalisation capability of existing enhancement methods when applied to unseen real-world dynamic low-light videos.

To ensure a consistent comparison, all models are evaluated on the same test sequences. The enhanced outputs are converted to RGB format and clipped to the valid intensity range. Since some methods generate outputs with spatial resolutions different from the input frames, the corresponding aligned normal-light references are resized or cropped, when required, to match the spatial resolution of each enhanced output before metric computation. This ensures that each enhanced output and its reference are compared over the same spatial dimensions. This resolution-normalisation step is applied only for metric computation and does not alter the original PiLLVD data. The dataset retains the original aligned paired frames, while metric evaluation uses spatially matched enhanced-reference pairs to account for methods with different output resolutions.

Quantitative scores are computed frame by frame between each enhanced output and its corresponding normal-light reference frame. The frame-level scores are then averaged over each video sequence and finally averaged across the complete test set. In addition to frame-level reconstruction quality, temporal consistency is evaluated across consecutive frames to assess whether each method preserves stable brightness and structural details over time. This is particularly important for PiLLVD, as the dataset contains dynamic scenes with moving humans, objects, and illumination variations.

\subsection{Evaluation Metrics}
\label{em}

We evaluate the enhanced videos using complementary fidelity, perceptual, and temporal stability metrics. Specifically, PSNR and SSIM \cite{ssim} are used to measure frame-level reconstruction quality with respect to the aligned normal-light references, where higher values indicate better fidelity. LPIPS \cite{lpips} is adopted to assess perceptual similarity using deep feature representations, where lower values indicate better perceptual quality. For temporal stability, we use a masked temporal warping error (mTWE), following \cite{blind_video_tc}. This metric compares consecutive enhanced frames after motion compensation, with lower values indicating better temporal consistency. Since the original implementation relies on an older FlowNet2-based environment, optical flow is estimated using RAFT \cite{raft}. Forward and backward optical flows are computed between consecutive low-light input frames and used to generate valid-region masks through forward--backward consistency checking. Pixels warped outside the image boundaries are also excluded. The enhanced frame at time $t$ is then warped towards frame $t+1$, and the average photometric difference is computed only over valid, non-occluded regions.

\begin{table*}[t]
\centering
\caption{Quantitative comparison of different enhancement models on outdoor PiLLVD sequences.}
\label{tab:QuanR_outdoor}

\renewcommand{\arraystretch}{1.4}
\resizebox{0.8\textwidth}{!}{%
\begin{tabular}{c|ccc|ccc|ccc|ccc}
\hline

\multirow{2}{*}{Model}
& \multicolumn{3}{c|}{PSNR $\uparrow$}
& \multicolumn{3}{c|}{SSIM $\uparrow$}
& \multicolumn{3}{c|}{LPIPS $\downarrow$}
& \multicolumn{3}{c}{mTWE $\downarrow$} \\
\cline{2-13}

& V2 & V5 & V7
& V2 & V5 & V7
& V2 & V5 & V7
& V2 & V5 & V7 \\
\hline

SDSDNet
& 17.133 & 13.456 & 19.737
& 0.726 & 0.575 & 0.582
& \textbf{0.154} & 0.250 & \textbf{0.190}
& 0.017 & 0.032 & 0.031 \\

DP3DF
& 17.956 & \textbf{21.180} & 22.133
& 0.699 & 0.693 & \textbf{0.756}
& 0.228 & 0.336 & 0.293
& \textbf{0.006} & 0.013 & \textbf{0.010} \\

LLVE-STCD
& 16.909 & 13.876 & 17.309
& \textbf{0.737} & 0.671 & 0.678
& 0.202 & 0.276 & 0.242
& 0.009 & \textbf{0.010} & 0.012 \\

StableLLVE
& 15.575 & 16.789 & 17.917
& 0.690 & \textbf{0.720} & 0.702
& 0.194 & \textbf{0.206} & 0.202
& 0.013 & 0.016 & 0.019 \\

SGLLIE/SGZ
& \textbf{18.415} & 20.900 & \textbf{22.738}
& 0.525 & 0.645 & 0.550
& 0.309 & 0.351 & 0.417
& 0.022 & 0.033 & 0.039 \\
\hline

\end{tabular}%
}
\end{table*}

\subsection{Quantitative Results}
\label{QR}

The quantitative evaluation is conducted on PiLLVD under the zero-shot protocol described above. The test set consists of four indoor videos, denoted as V4, V5, V14, and V17, and three outdoor videos, denoted as V2, V5, and V7. These sequences were selected to cover a broad range of the dataset variation, including indoor and outdoor environments, different illumination levels, diverse scene structures, and both human and object motion. Together, these videos cover different acquisition conditions and motion patterns, including object motion, repeated motion, human motion, and less regular dynamic activity. 
For the indoor sequences, V4 includes an object moving back and forth, V5 shows cups being lifted, moved, and placed on a table, V14 contains a person standing up and sitting down, and V17 shows a person wiping a whiteboard. For the outdoor sequences, V2 includes repeated arm movement, V5 contains a person walking, and V7 shows a person climbing a wall. This selection demonstrates the ability of PiLLVD to assess both frame-level enhancement quality and temporal stability under diverse dynamic conditions.

Tables~\ref{tab:QuanR_indoor} and \ref {tab:QuanR_outdoor} report the quantitative results for the selected indoor and outdoor test sequences, respectively. The best result for each video and metric is highlighted in bold. As shown in the table, performance varies substantially across models, metrics, and scene types. For the indoor sequences, LLVE-STCD achieves the strongest overall performance. It obtains the best SSIM values on all indoor videos, the best PSNR on V5, V14, and V17, and the lowest LPIPS on V5, V14, and V17. This suggests that LLVE-STCD is particularly effective for indoor dynamic scenes, where the motion is moderate and the background structure is relatively stable. StableLLVE achieves the lowest LPIPS on V4, indicating better perceptual similarity in this case, even though it does not obtain the highest PSNR or SSIM. SGLLIE/SGZ achieves the highest PSNR on V4, but its SSIM, LPIPS, and mTWE values are less competitive, showing that high PSNR alone does not necessarily correspond to better structural preservation, perceptual quality, or temporal stability.
According to mTWE, DP3DF performs strongly across both
indoor and outdoor sequences. It achieves the lowest mTWE
on indoor sequences V5, V14, and V17, and outdoor
sequences V2 and V7. LLVE-STCD achieves the lowest
mTWE on indoor sequence V4 and outdoor sequence V5.
Overall, DP3DF demonstrates stronger temporal consistency
across the benchmark. This highlights the importance of
evaluating video enhancement methods using temporal
metrics, since visually plausible individual frames may
still suffer from flickering or unstable brightness over time.

For the outdoor sequences, the results are more varied. SGLLIE/SGZ achieves the highest PSNR on V2 and V7, while DP3DF achieves the highest PSNR on V5 and the best SSIM on V7. LLVE-STCD obtains the best SSIM on V2, and StableLLVE achieves the best SSIM on V5. In terms of LPIPS, SDSDNet performs best on V2 and V7, while StableLLVE performs best on V5. These mixed results suggest that outdoor scenes are more challenging for the tested methods, likely due to stronger illumination variation, larger scene depth, and more complex motion.

Overall, no single method performs best across all metrics and all videos. Methods that achieve high frame-level fidelity do not always provide the best perceptual quality or temporal stability. This confirms the importance of evaluating low-light video enhancement using complementary metrics, particularly for dynamic scenes with different motion patterns and illumination conditions. The results also demonstrate that PiLLVD provides a challenging benchmark for assessing the zero-shot generalisation ability of existing enhancement models under realistic indoor and outdoor low-light video conditions.

\subsection{Qualitative Comparison}
\label{QC}

To complement the quantitative evaluation, qualitative comparisons are presented for representative indoor and outdoor PiLLVD sequences. Unlike single-frame comparisons, each example is shown using three consecutive frames to better reflect the video nature of the proposed dataset and to allow visual inspection of temporal behaviour across time. The comparison includes the low-light input, the outputs of the benchmarked enhancement methods, and the aligned normal-light reference.

\begin{figure*}
\centering

{%
\setlength{\tabcolsep}{0pt}
\renewcommand{\arraystretch}{1.1}

\begin{tabular}{
>{\centering\arraybackslash}m{1.9cm}
@{}>{\centering\arraybackslash}m{0.25\textwidth}@{}
@{\hspace{2mm}}
@{}>{\centering\arraybackslash}m{0.25\textwidth}@{}
@{\hspace{2mm}}
@{}>{\centering\arraybackslash}m{0.25\textwidth}@{}}

{} & \textbf{Frame $t$} & \textbf{Frame $t+1$} & \textbf{Frame $t+2$}\\

Input
& \includegraphics[width=\linewidth]{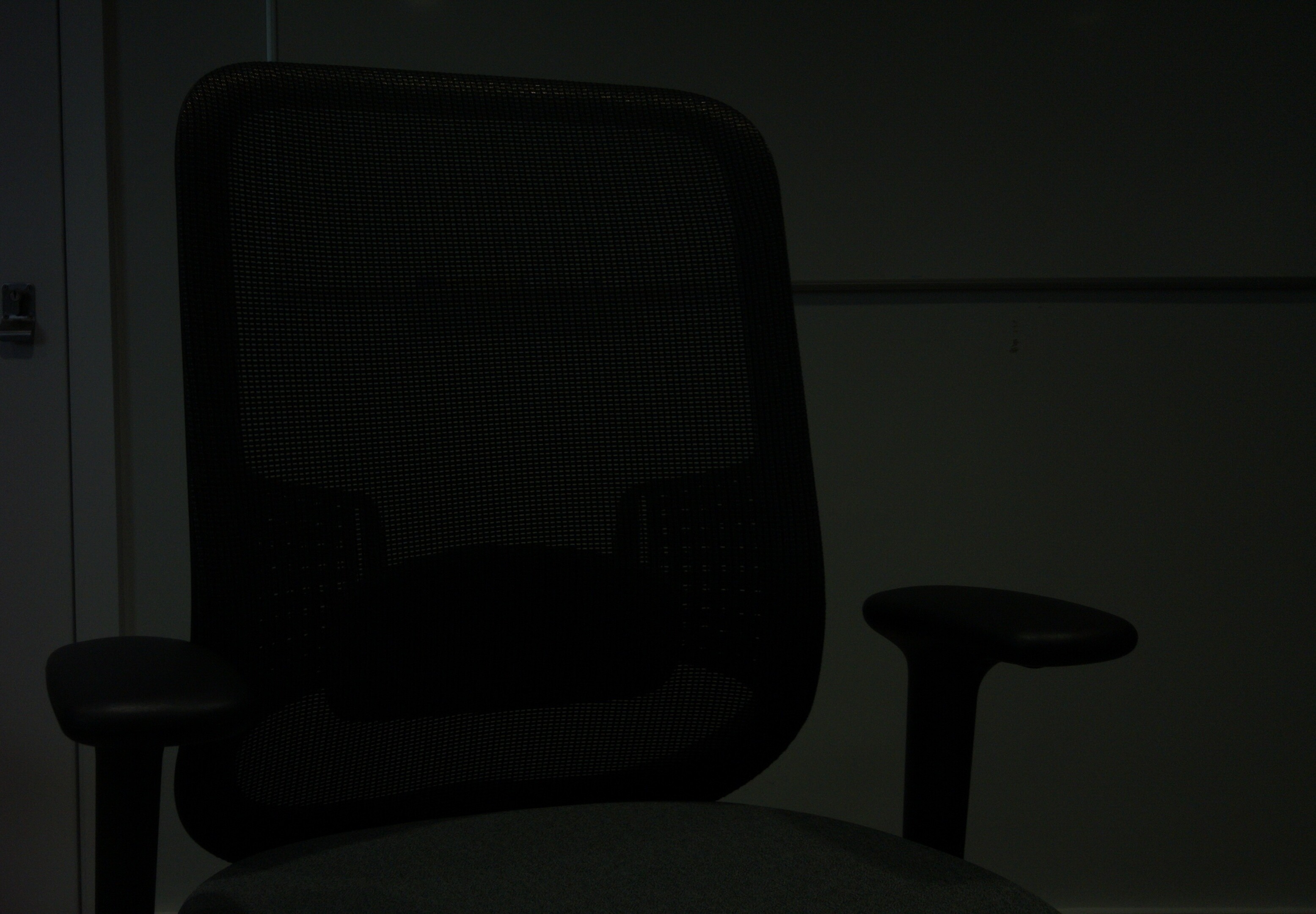}
& \includegraphics[width=\linewidth]{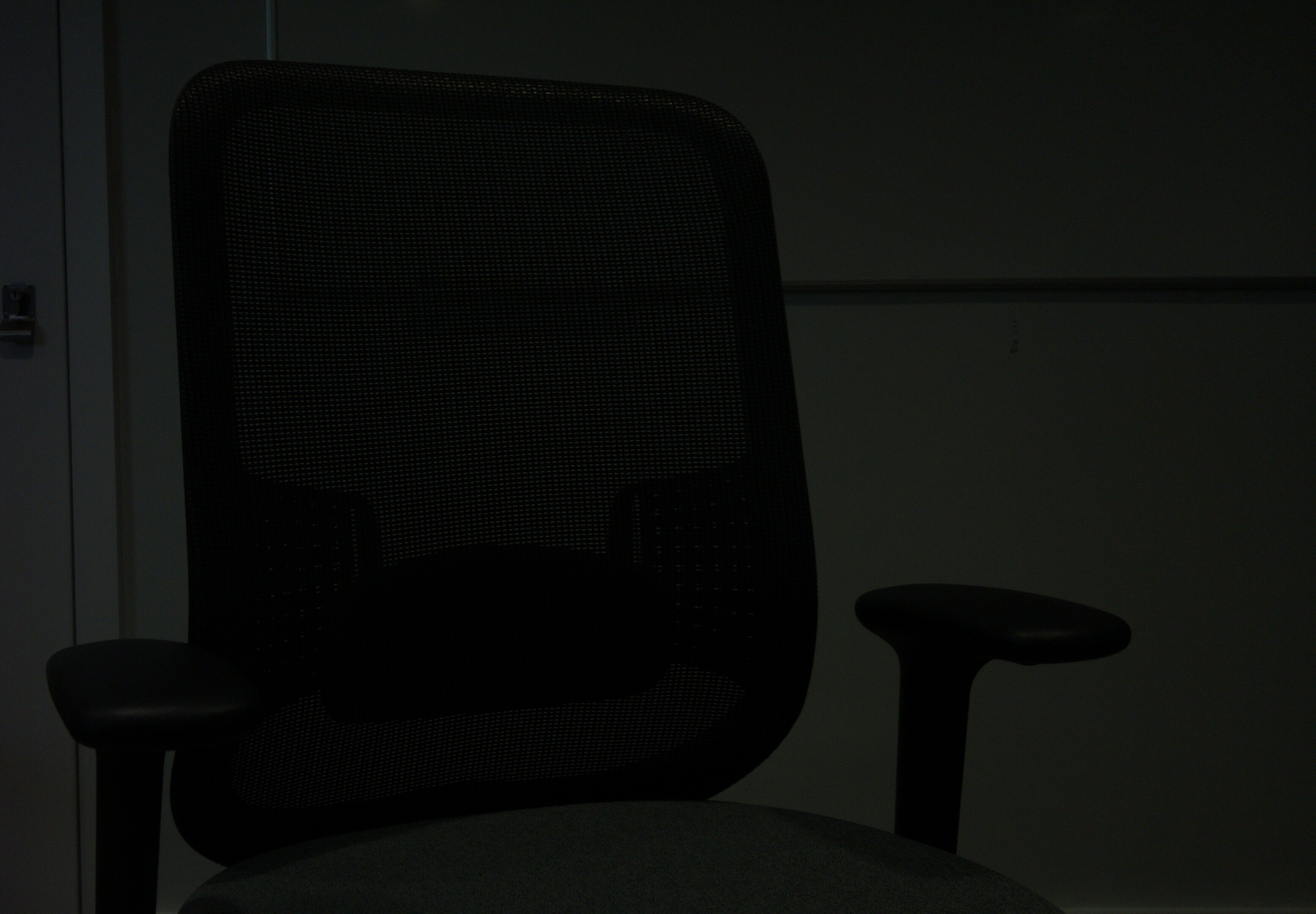}
& \includegraphics[width=\linewidth]{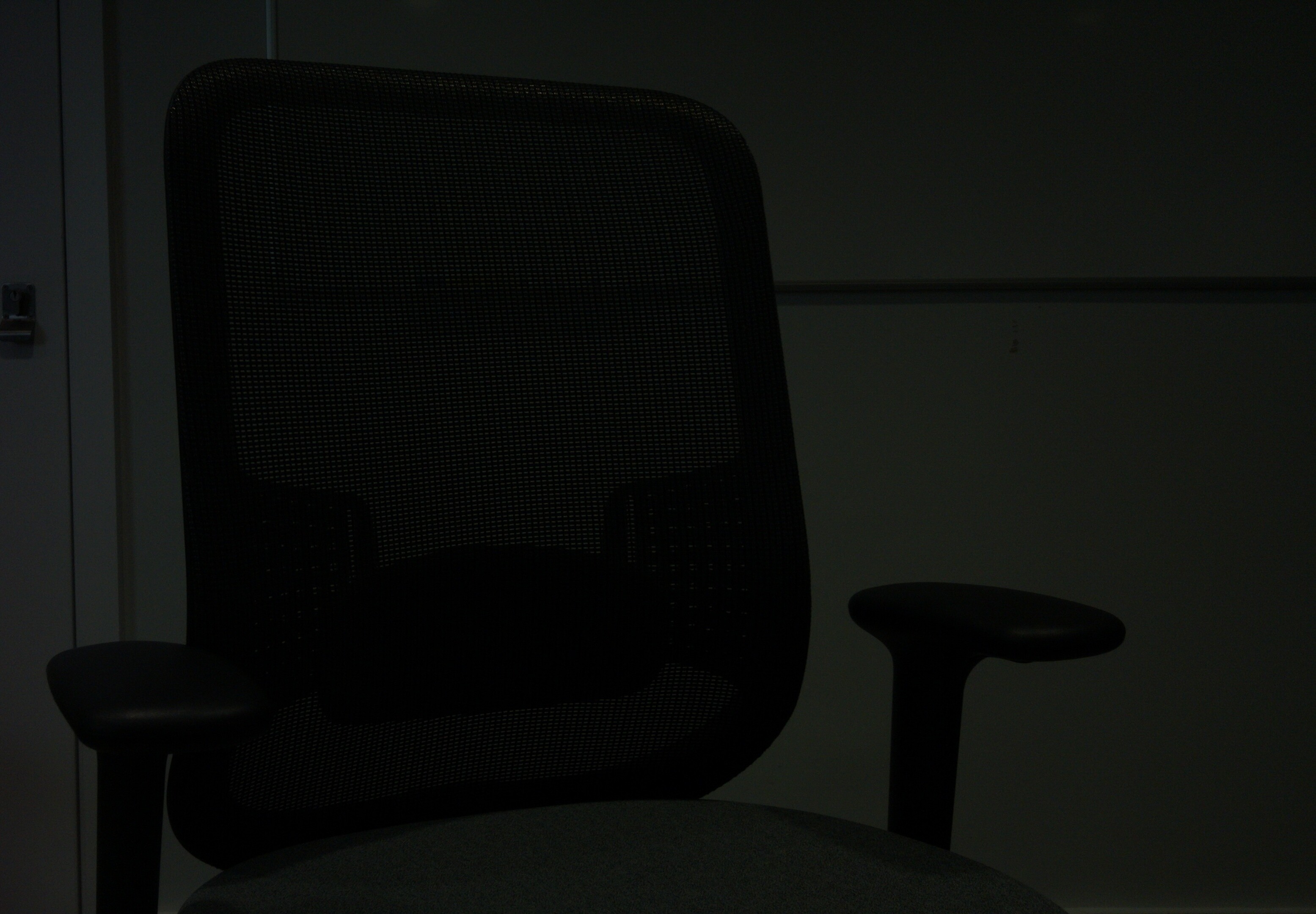} \\

SDSDNet
& \includegraphics[width=\linewidth]{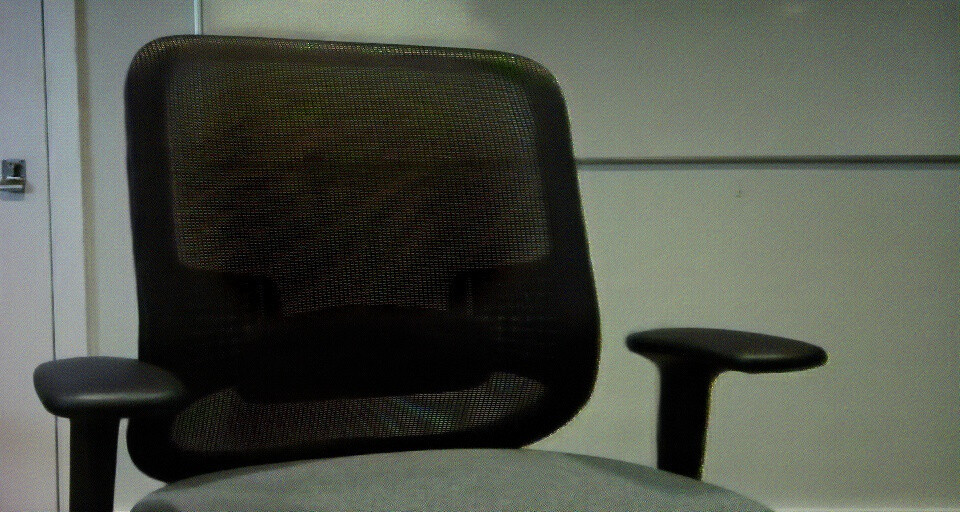}
& \includegraphics[width=\linewidth]{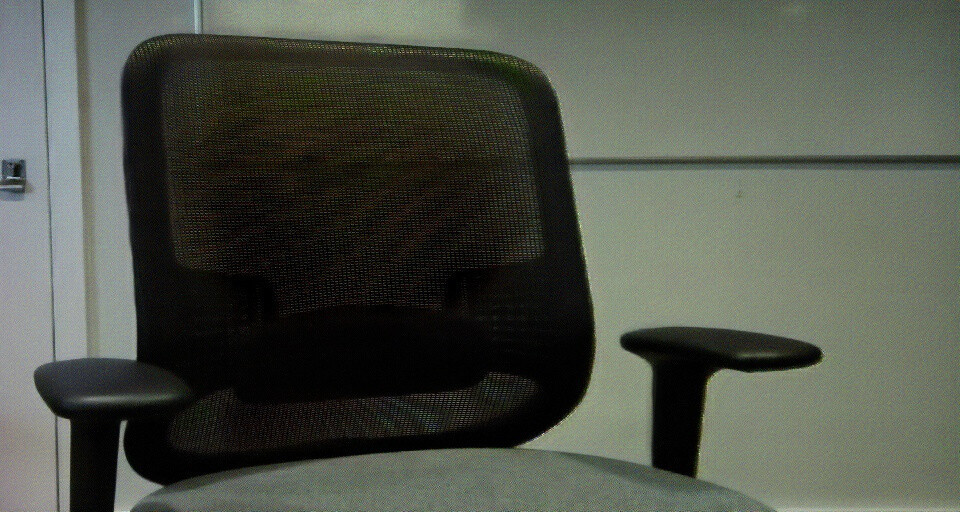}
& \includegraphics[width=\linewidth]{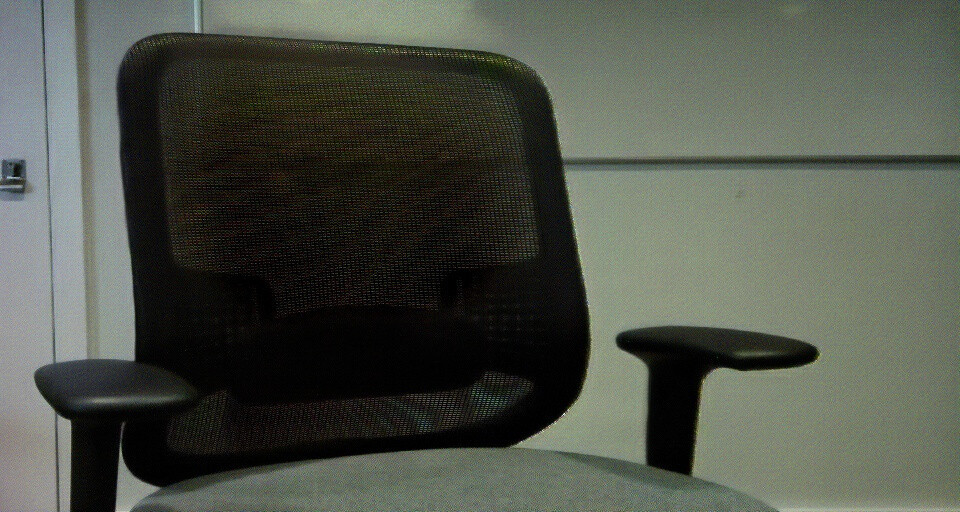} \\

DP3DF
& \includegraphics[width=\linewidth]{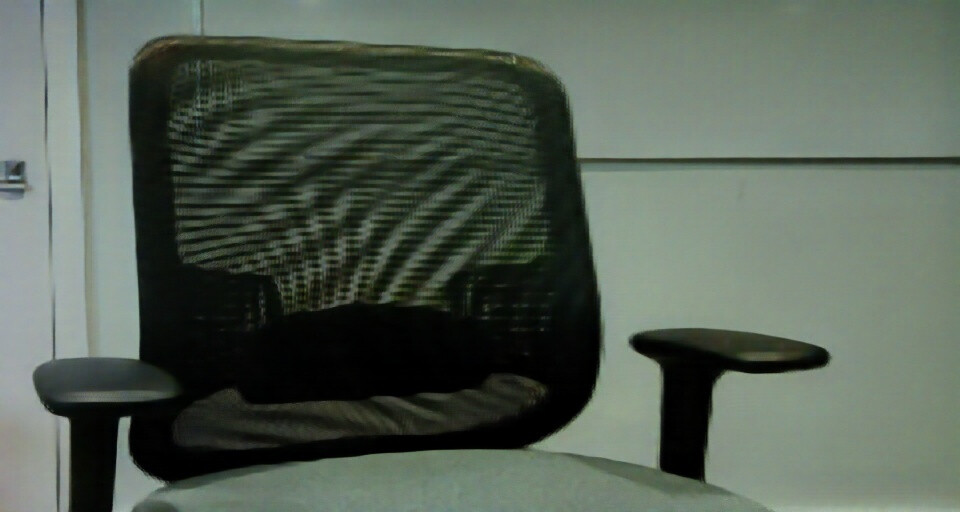}
& \includegraphics[width=\linewidth]{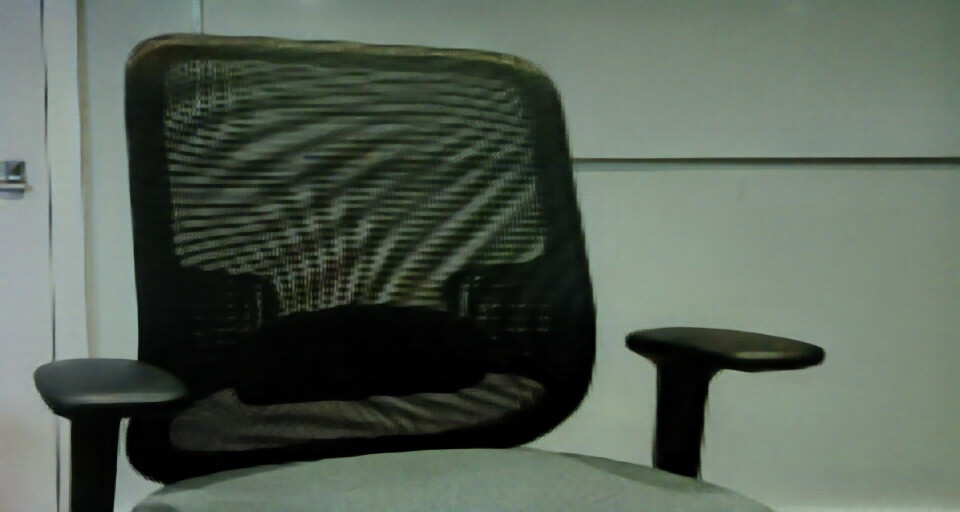}
& \includegraphics[width=\linewidth]{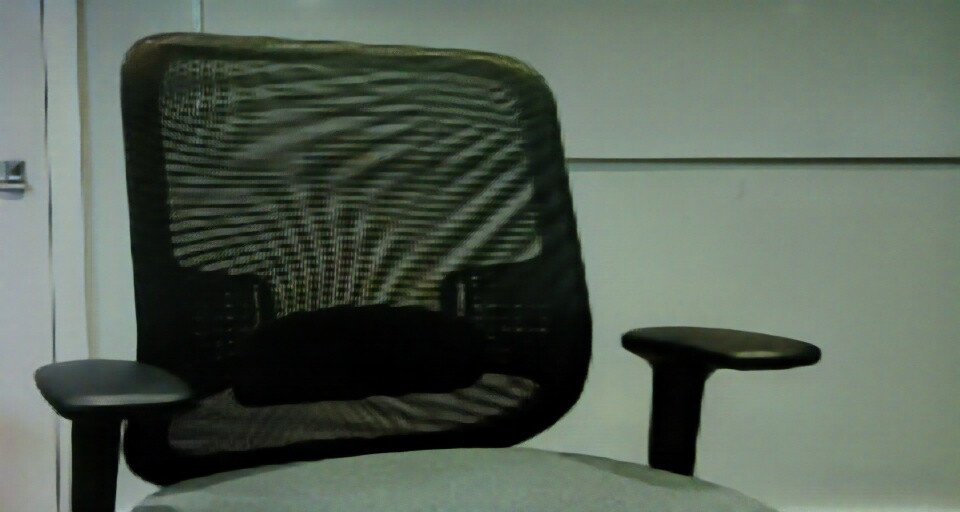} \\

LLVE-STCD
& \includegraphics[width=\linewidth]{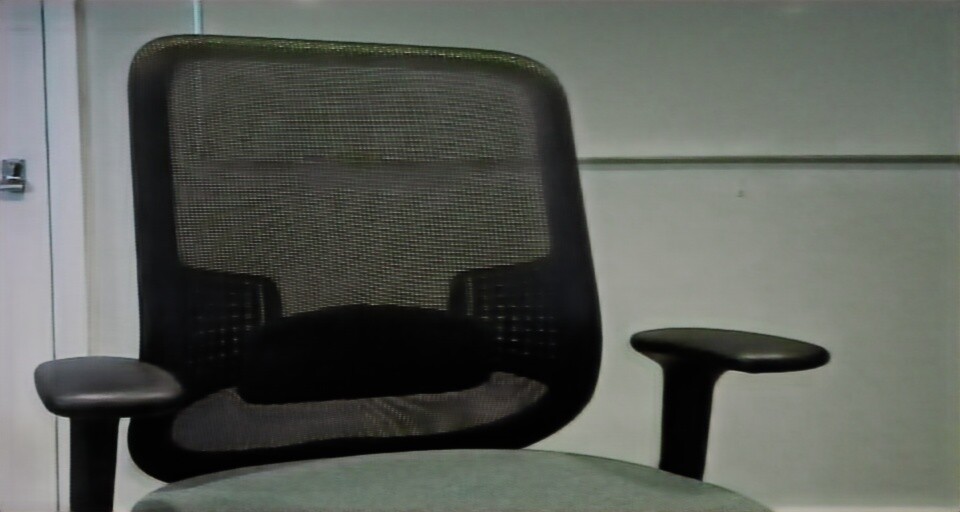}
& \includegraphics[width=\linewidth]{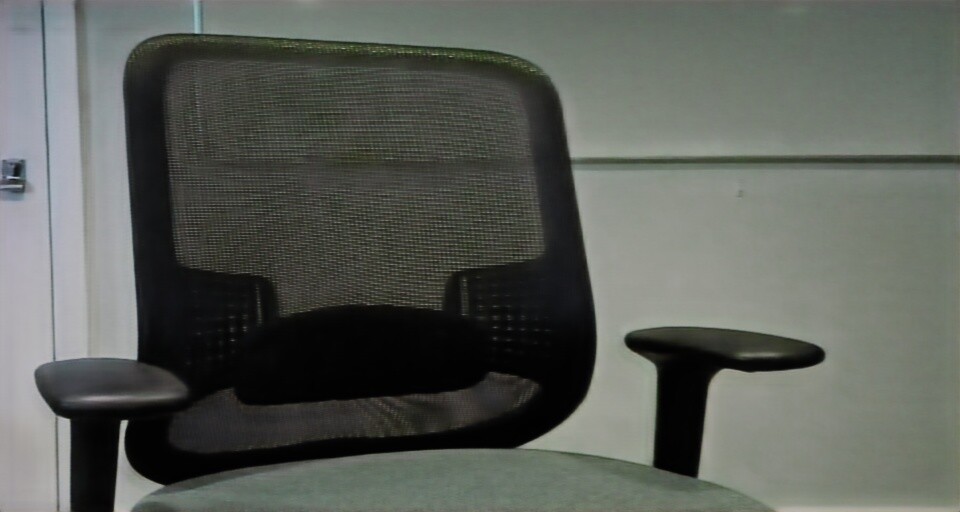}
& \includegraphics[width=\linewidth]{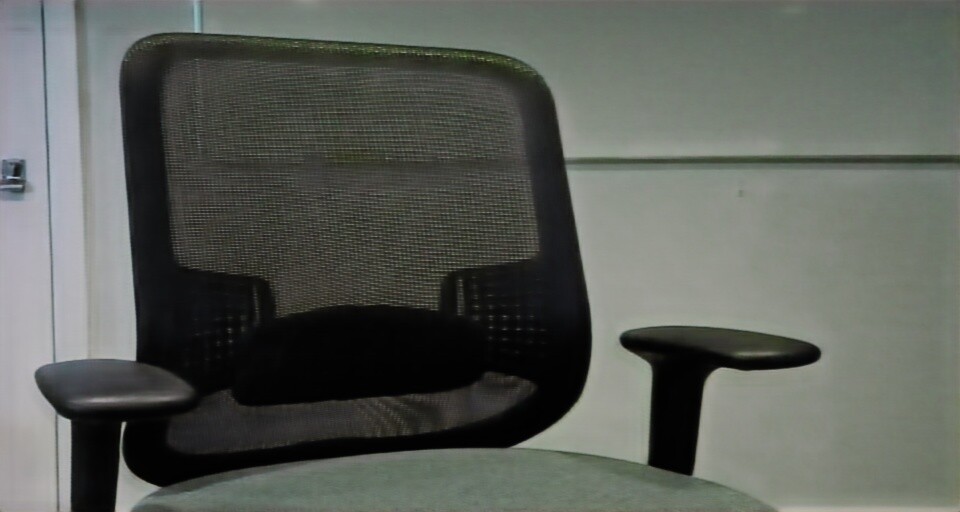} \\

StableLLVE
& \includegraphics[width=\linewidth]{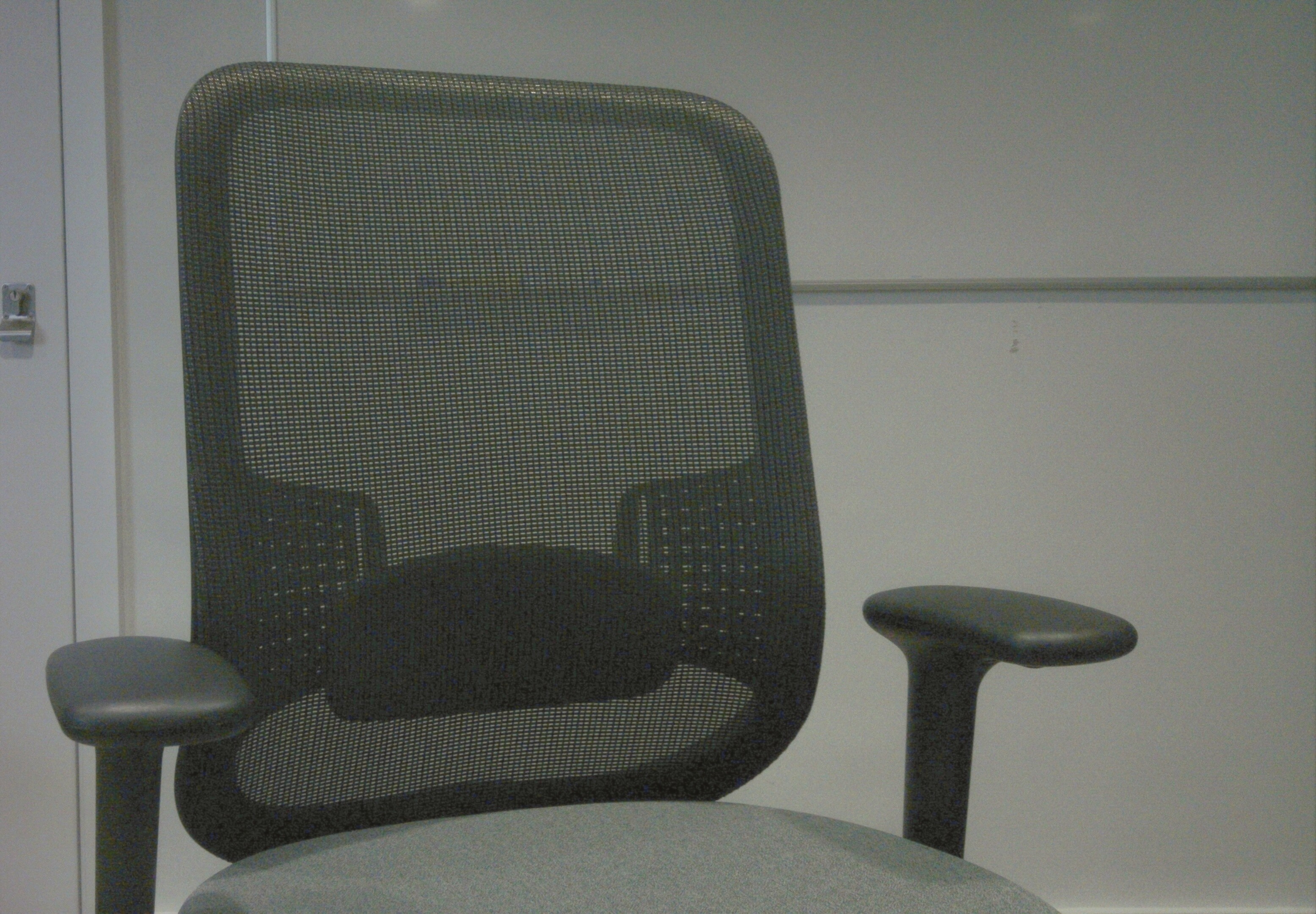}
& \includegraphics[width=\linewidth]{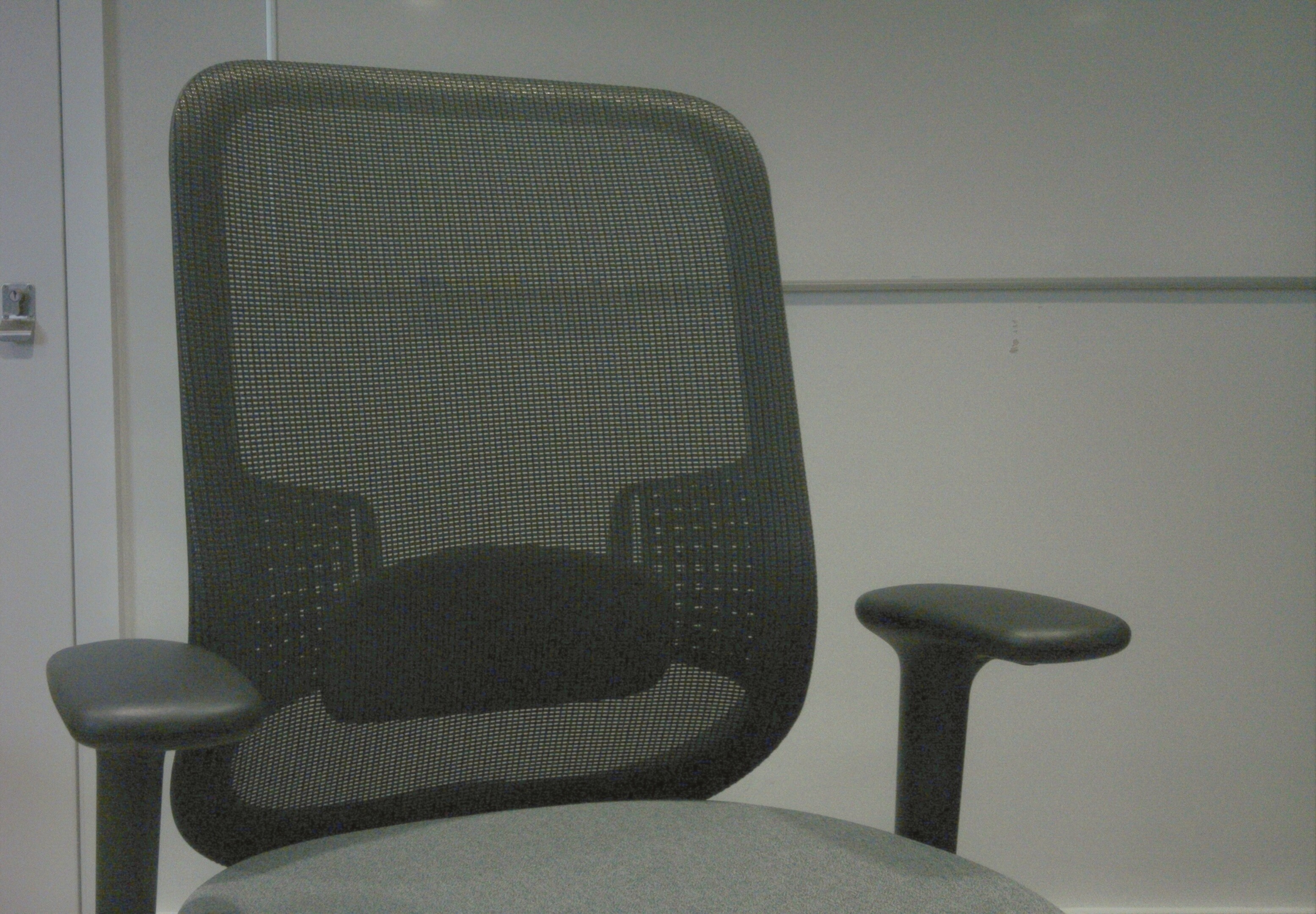}
& \includegraphics[width=\linewidth]{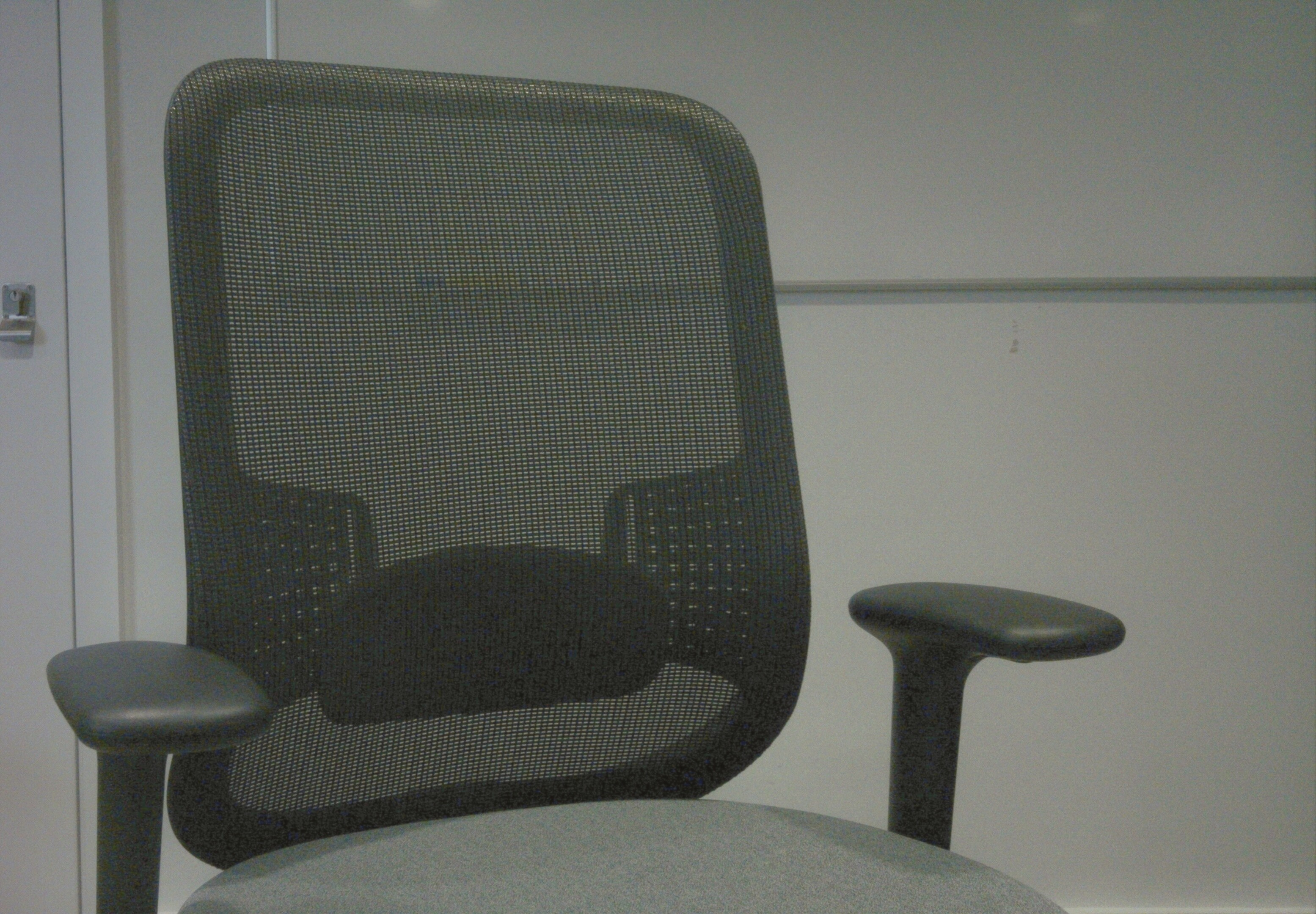} \\

\makecell{SGLLIE/\\SGZ}
& \includegraphics[width=\linewidth]{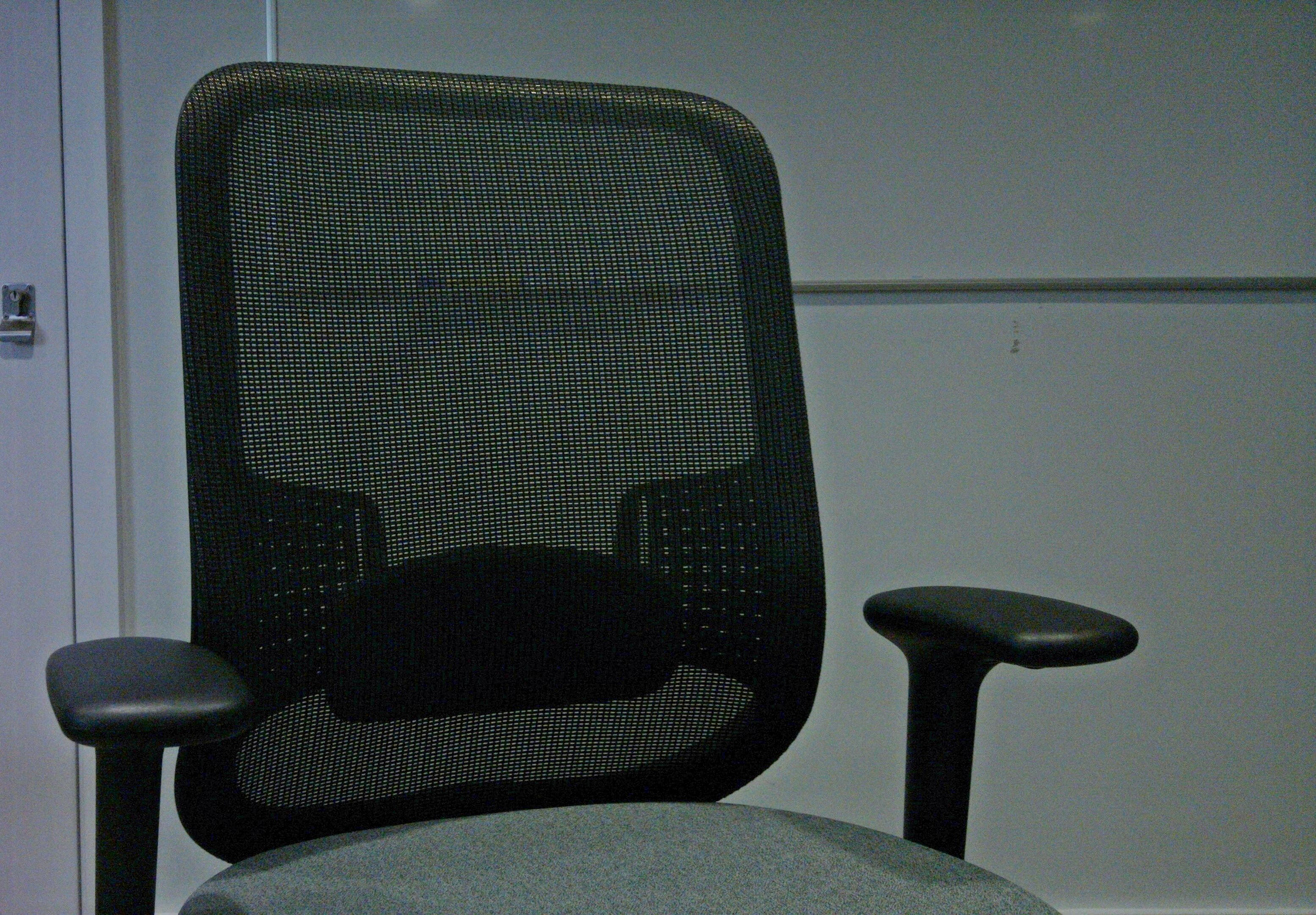}
& \includegraphics[width=\linewidth]{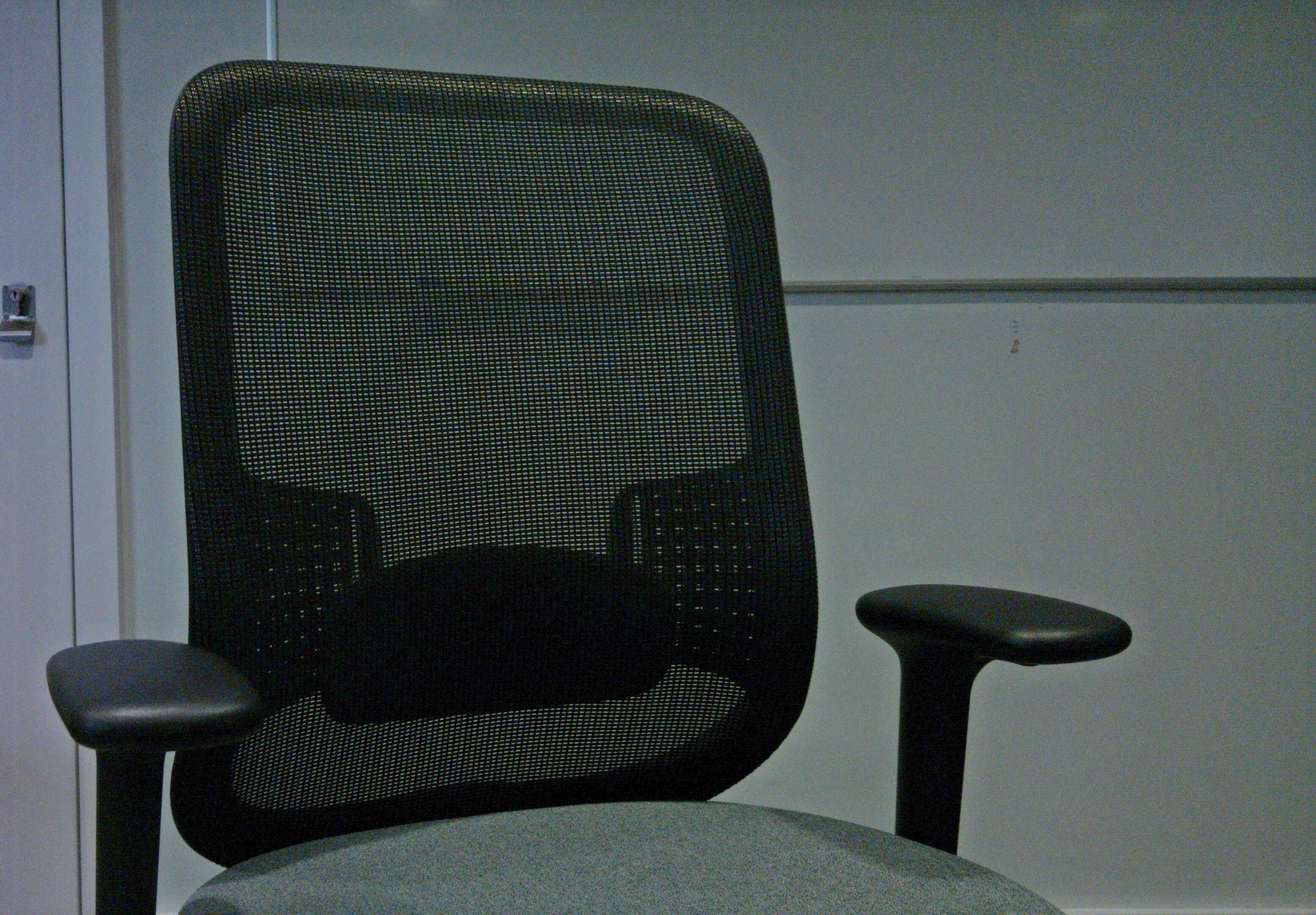}
& \includegraphics[width=\linewidth]{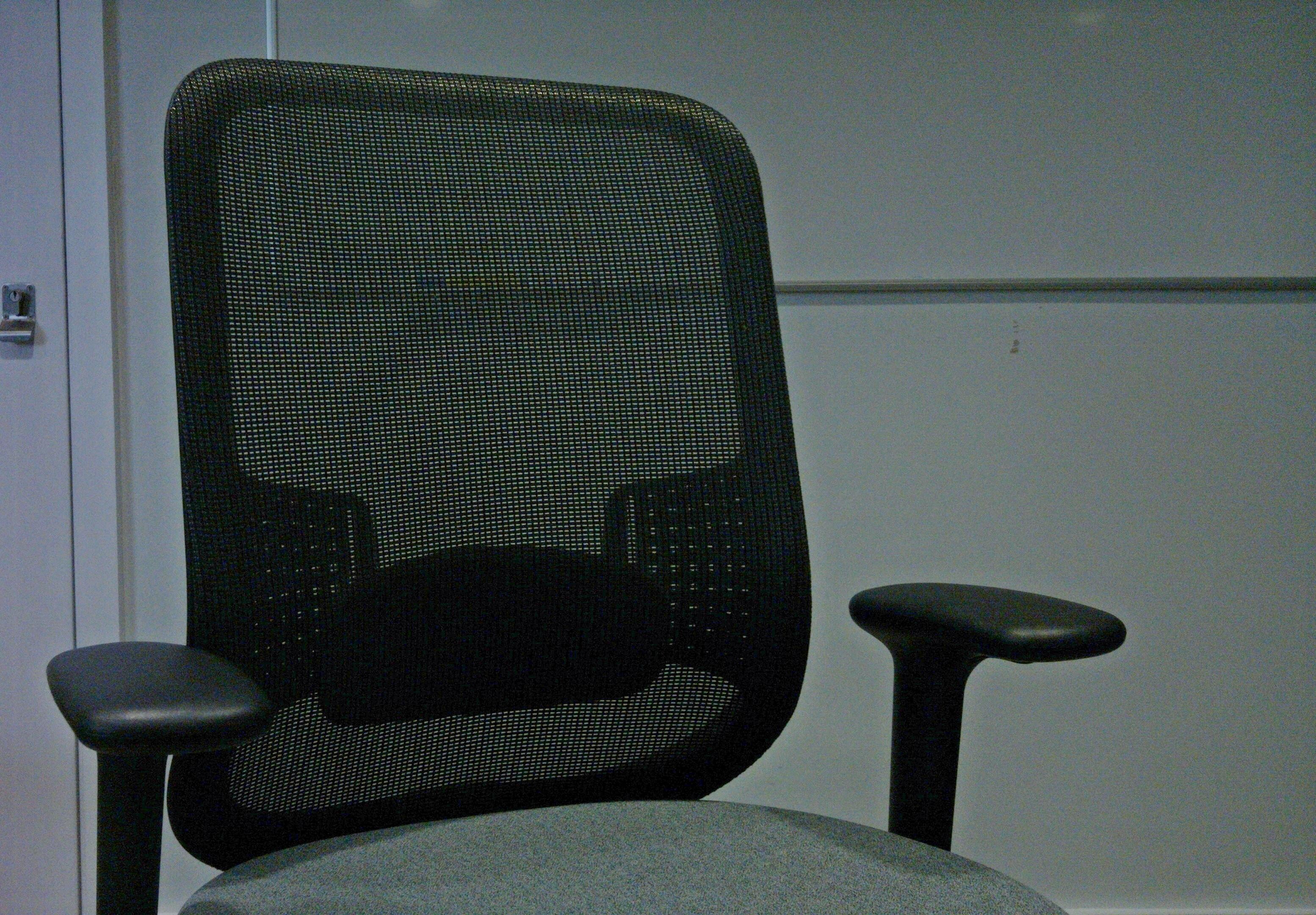} \\

Reference
& \includegraphics[width=\linewidth]{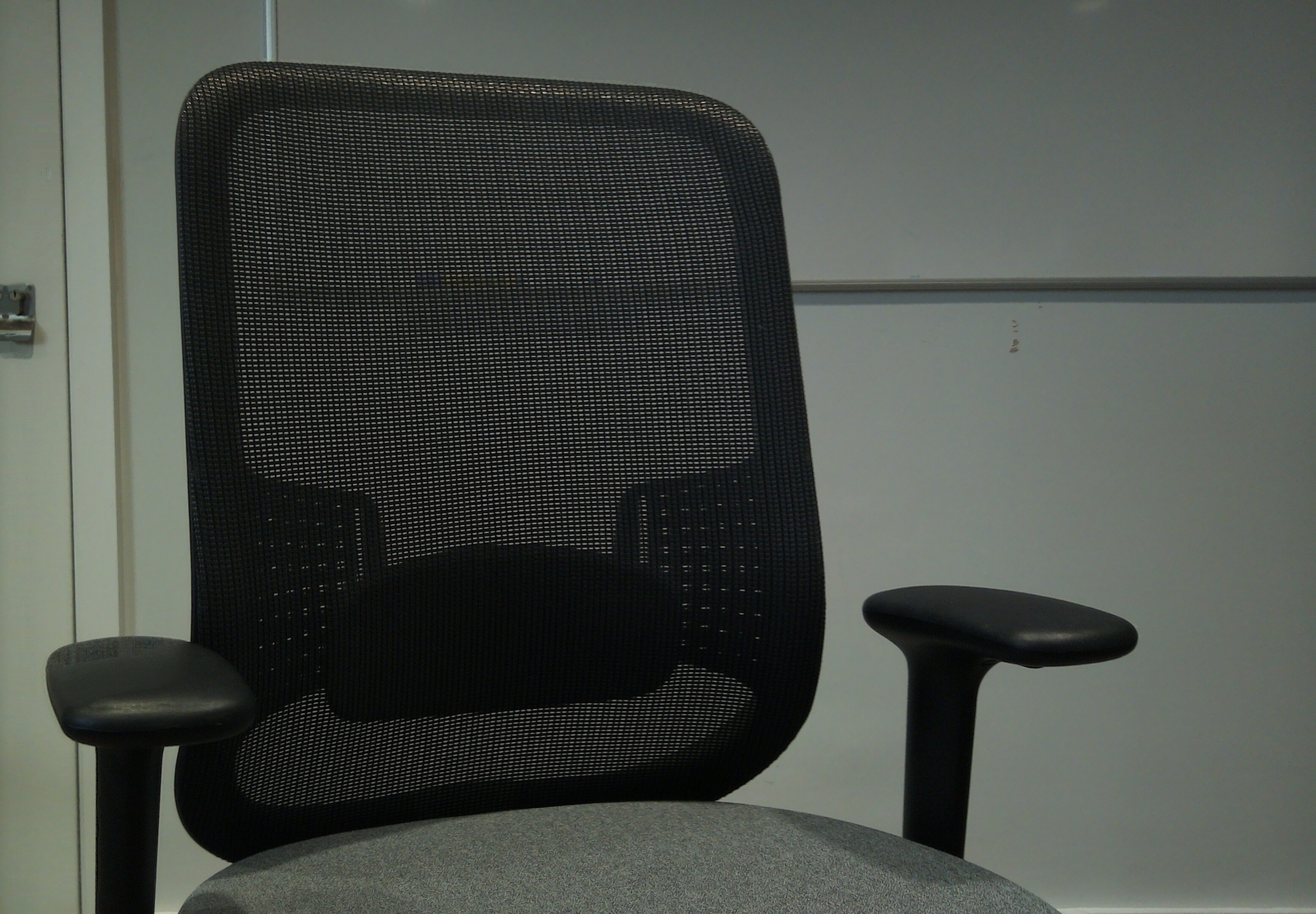}
& \includegraphics[width=\linewidth]{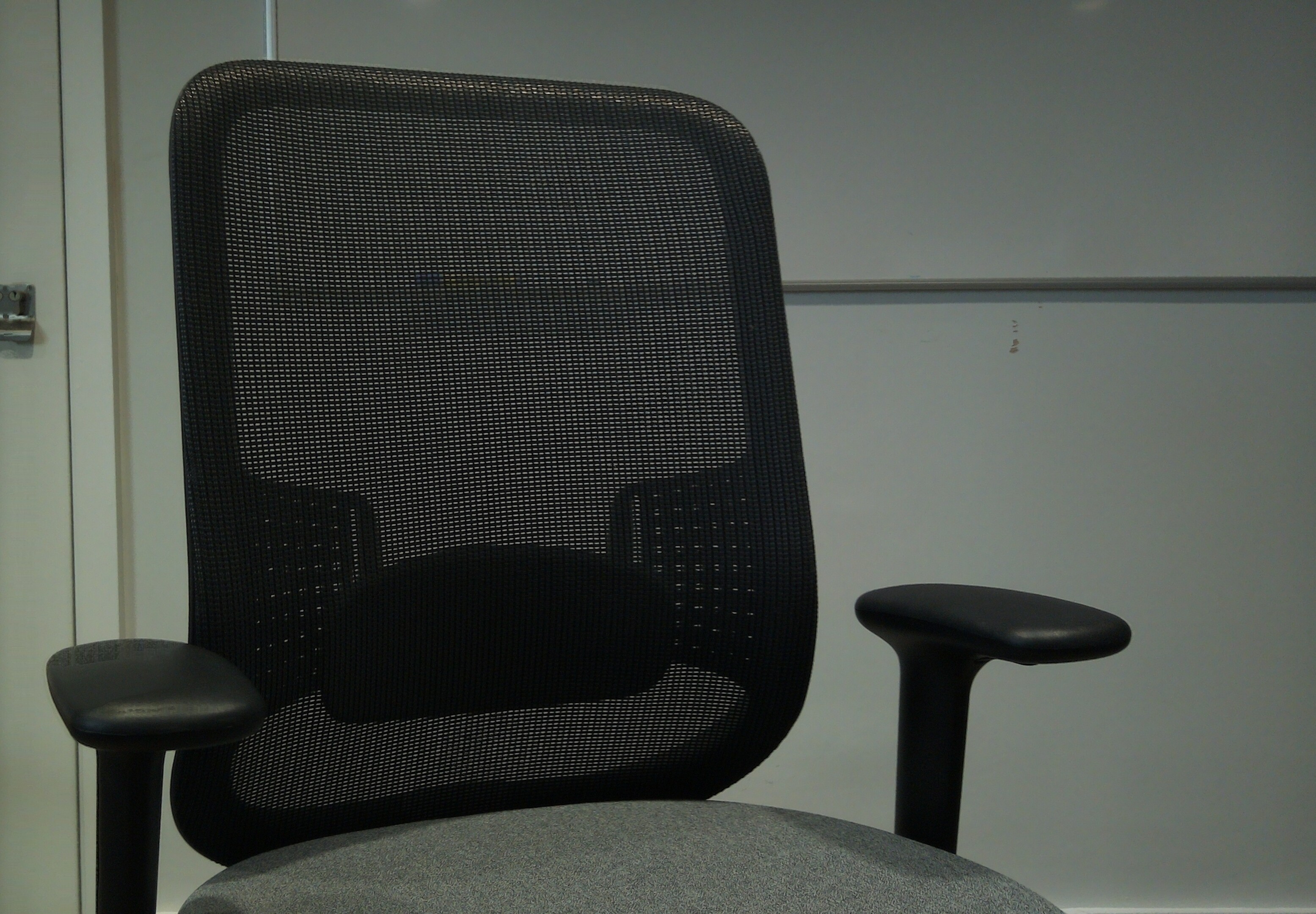}
& \includegraphics[width=\linewidth]{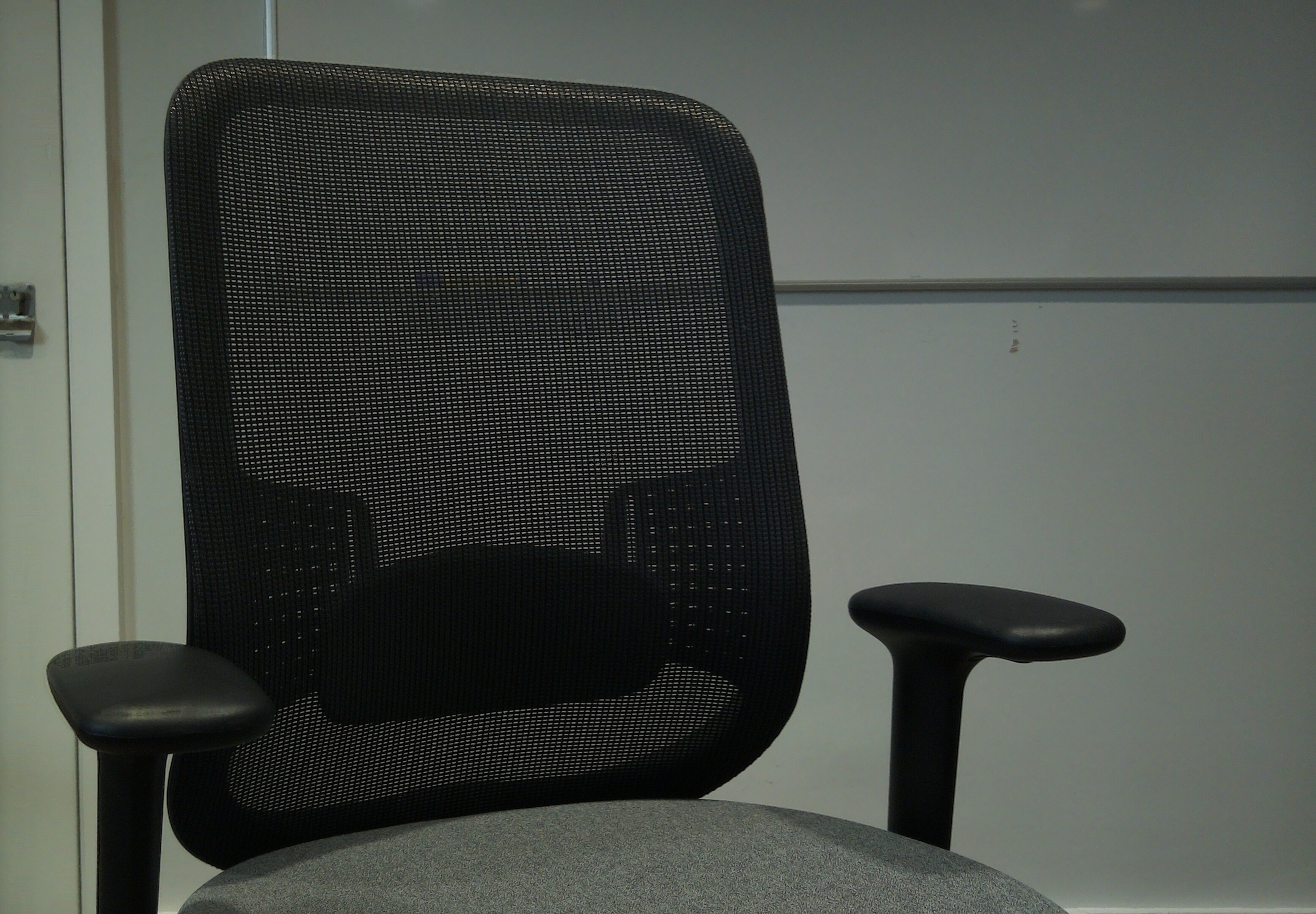} \\

\end{tabular}
}%

\caption{Qualitative comparison on three consecutive frames from a representative indoor PiLLVD sequence. The comparison includes the low-light input, outputs of the benchmarked enhancement methods, and the aligned normal-light reference.}
\label{fig:qual_indoor}
\end{figure*}

Figure~\ref{fig:qual_indoor} shows the indoor comparison. The low-light input is strongly under-exposed, and fine structural details of the chair, wall, and surrounding scene are difficult to distinguish. Compared with the aligned normal-light reference, the enhanced outputs differ noticeably in brightness correction, contrast, colour reproduction, noise suppression, and detail preservation. 
SDSDNet improves the global brightness and reveals the main chair structure, but the output shows noticeable loss of fine detail and noise, particularly on the wall and chair surface.
DP3DF makes the chair mesh highly visible; however, the mesh texture appears over-amplified and less natural than the reference, with a noticeable greenish colour shift. LLVE-STCD provides a more balanced enhancement in terms of global brightness and contrast, although some fine mesh details remain smoothed or less faithful to the reference. This visual behaviour is consistent with its strong indoor SSIM and LPIPS performance reported in Table~\ref{tab:QuanR_indoor}. StableLLVE preserves the overall chair structure and produces a smoother result, but the output appears hazy and lower in contrast, with weakened high-frequency texture. SGLLIE/SGZ preserves the chair mesh structure reasonably well in this example, but its brightness and colour tone still differ from the aligned normal-light reference. These observations show that visually plausible enhancement does not depend only on brightness correction; different methods trade off noise suppression, texture preservation, colour fidelity, and structural consistency.

The consecutive indoor frames also show differences in temporal behaviour. Since the camera and scene content remain relatively stable across the three frames, a temporally consistent method should maintain similar brightness, colour, and local texture over time. DP3DF and LLVE-STCD show relatively stable frame-to-frame appearance, which agrees with their strong mTWE values on the indoor sequences. In contrast, methods that strongly alter brightness or texture may produce visually noticeable variations across frames, particularly around detailed regions such as the chair mesh and background wall. This demonstrates that temporal stability is not only related to object motion, but also to the consistency of illumination and texture reconstruction across adjacent frames.

\begin{figure*}
\centering

{%
\setlength{\tabcolsep}{2pt}
\renewcommand{\arraystretch}{1.1}

\begin{tabular}{
>{\centering\arraybackslash}m{1.9cm}
@{}>{\centering\arraybackslash}m{0.25\textwidth}@{}
@{\hspace{2mm}} 
@{}>{\centering\arraybackslash}m{0.25\textwidth}@{}
@{\hspace{2mm}}
@{}>{\centering\arraybackslash}m{0.25\textwidth}@{}}

{} & \textbf{Frame $t$} & \textbf{Frame $t+1$} & \textbf{Frame $t+2$}\\

Input
& \includegraphics[width=\linewidth]{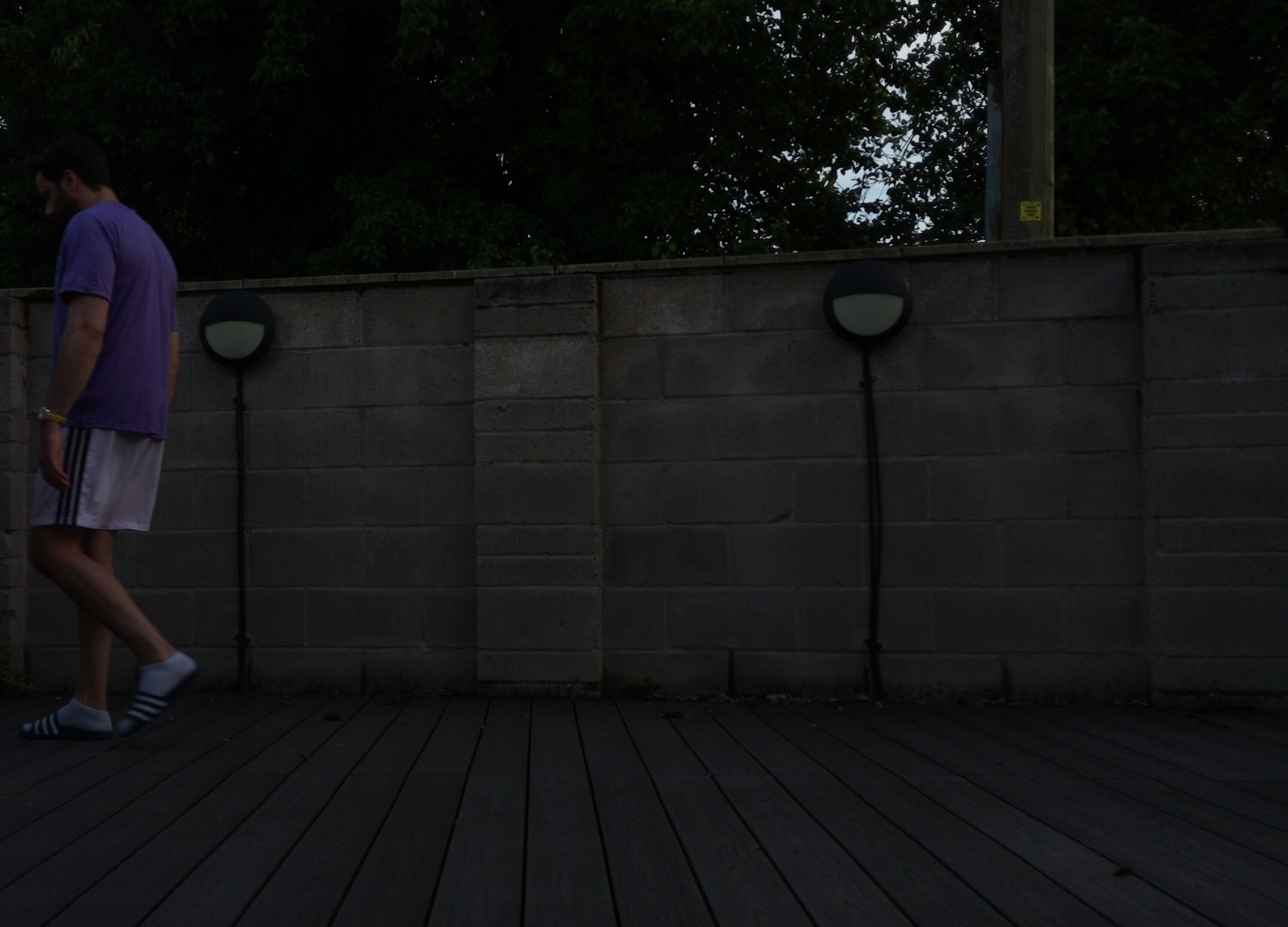}
& \includegraphics[width=\linewidth]{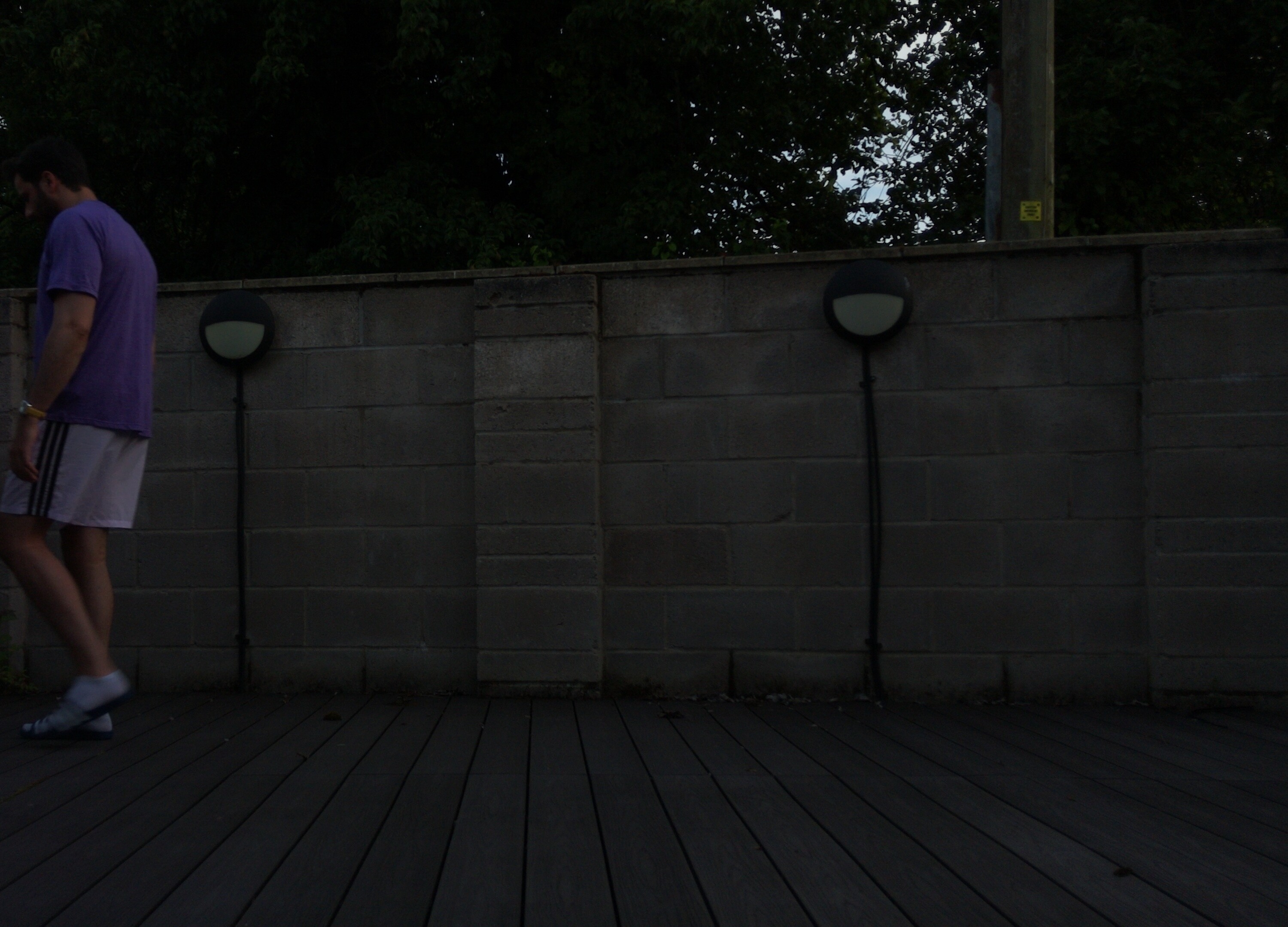}
& \includegraphics[width=\linewidth]{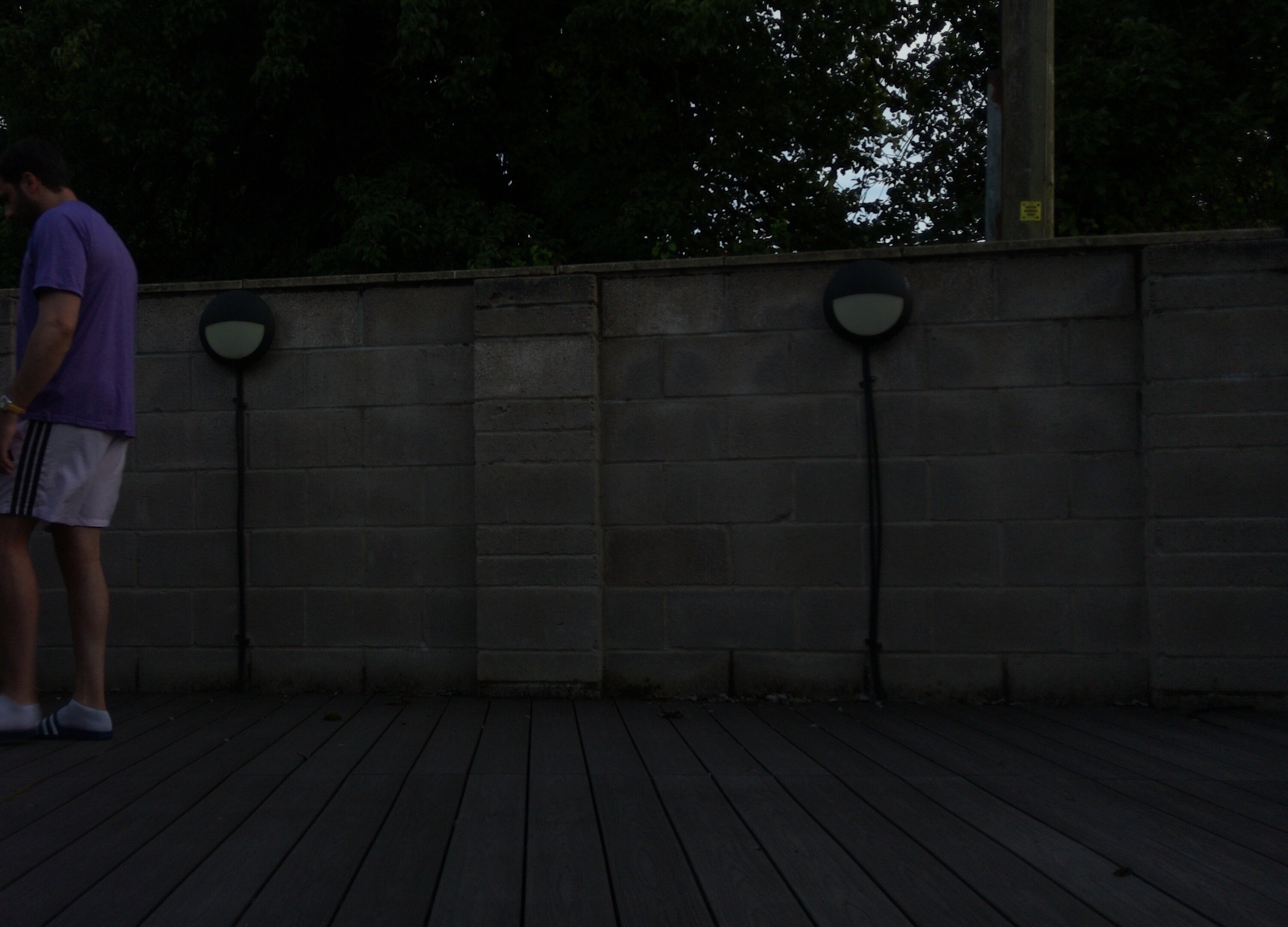} \\

SDSDNet
& \includegraphics[width=\linewidth]{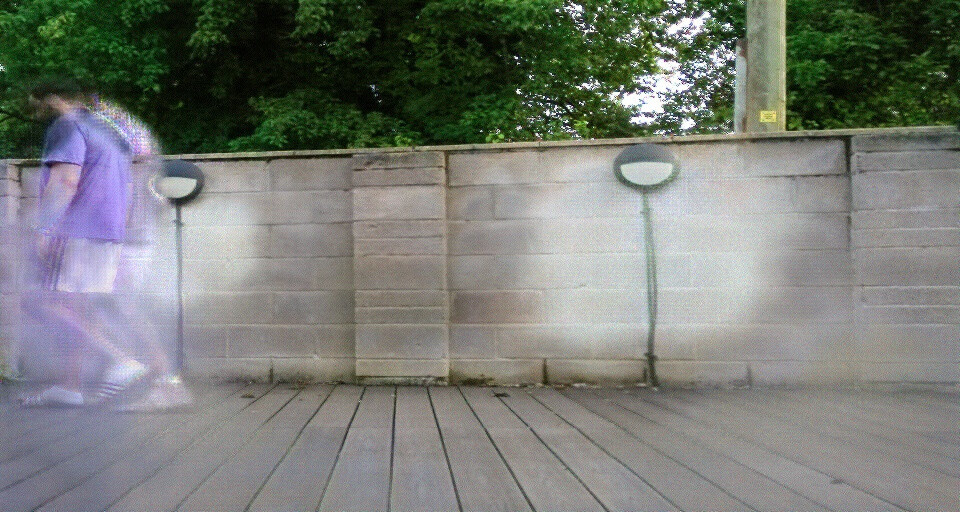}
& \includegraphics[width=\linewidth]{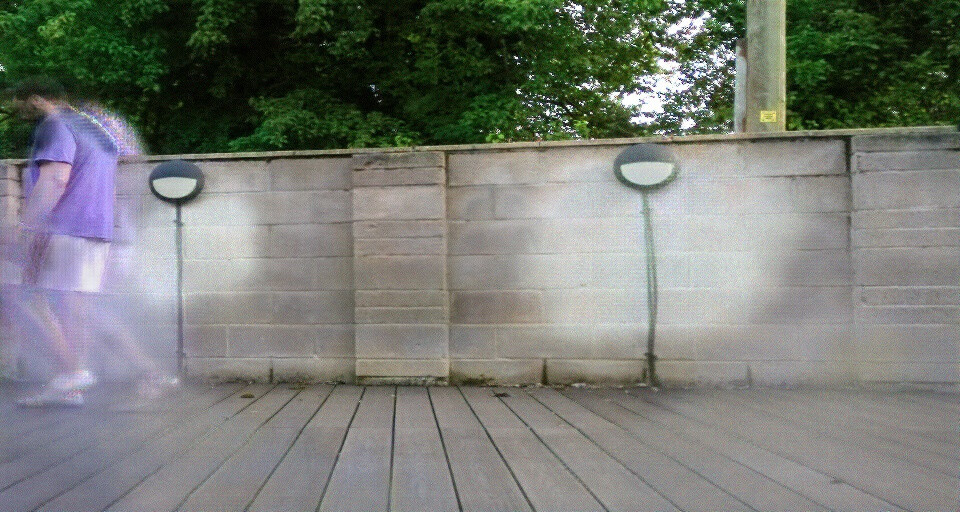}
& \includegraphics[width=\linewidth]{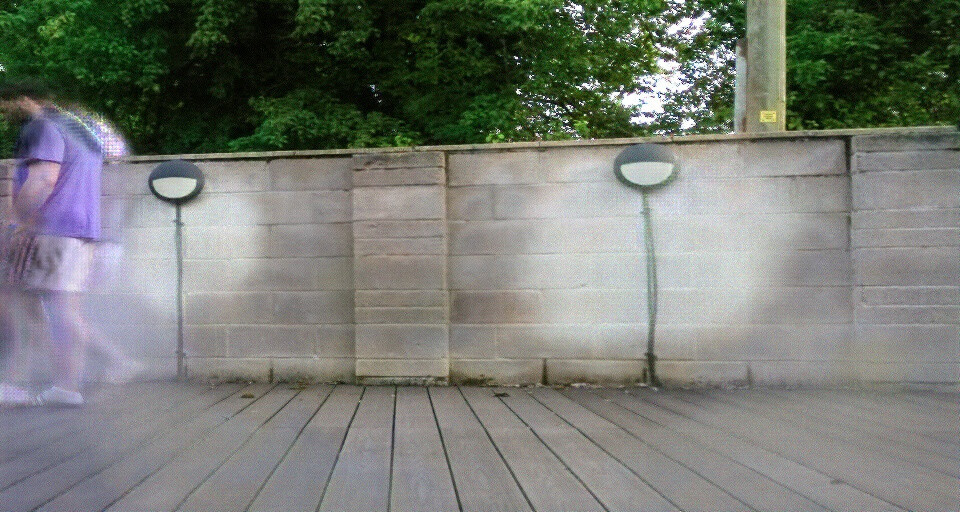} \\

DP3DF
& \includegraphics[width=\linewidth]{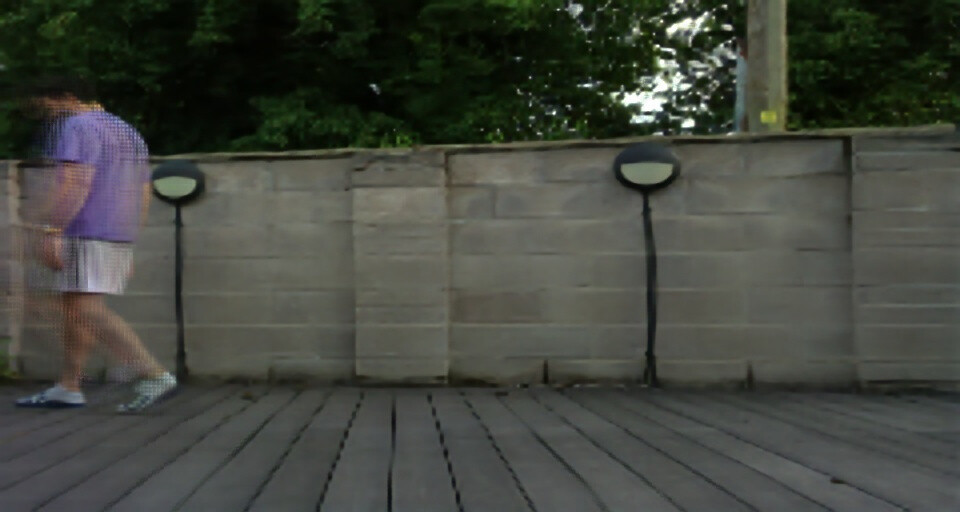}
& \includegraphics[width=\linewidth]{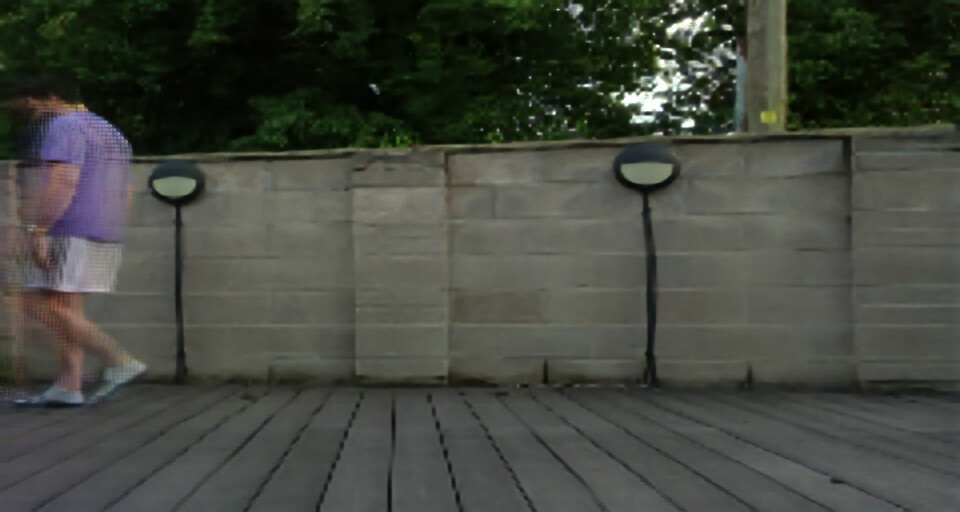}
& \includegraphics[width=\linewidth]{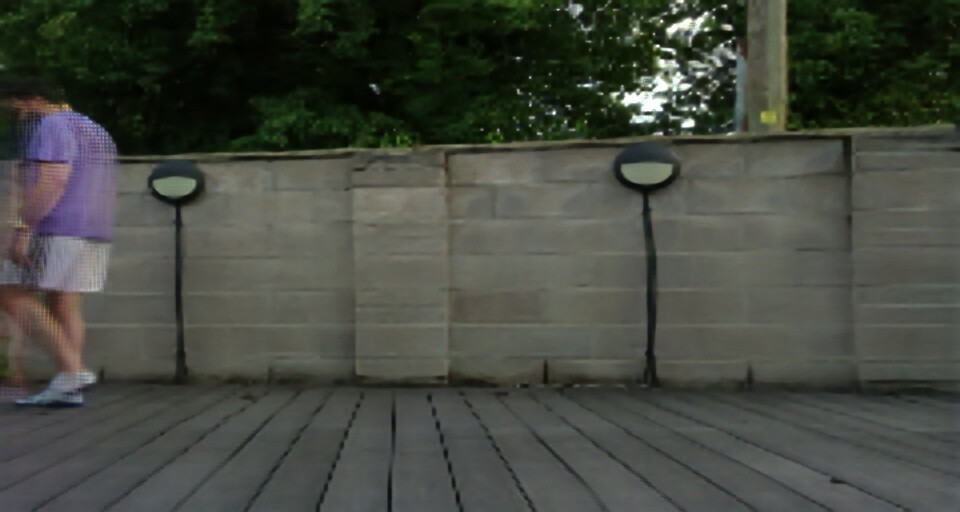} \\

LLVE-STCD
& \includegraphics[width=\linewidth]{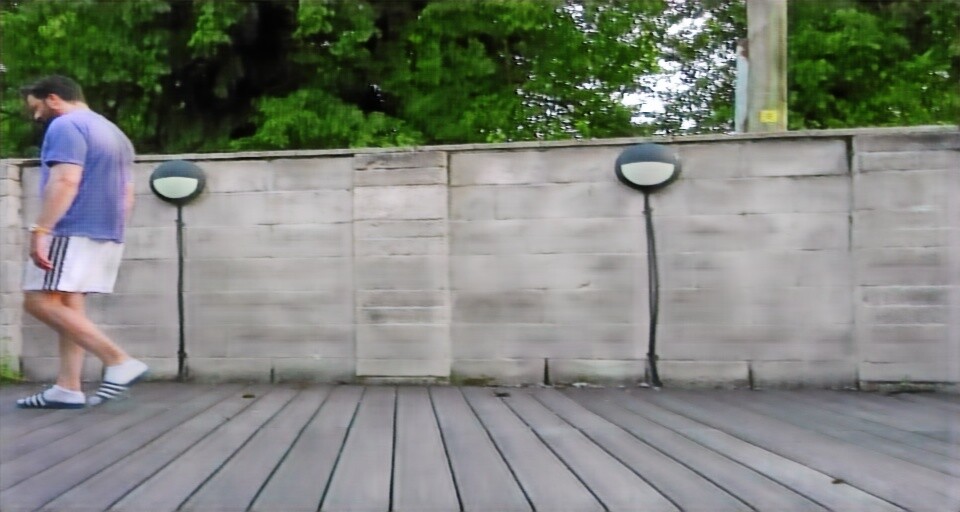}
& \includegraphics[width=\linewidth]{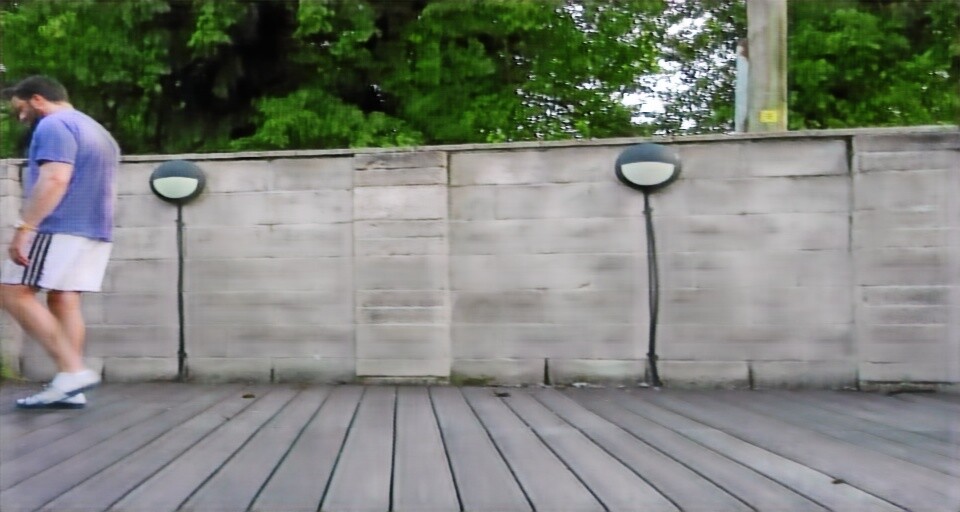}
& \includegraphics[width=\linewidth]{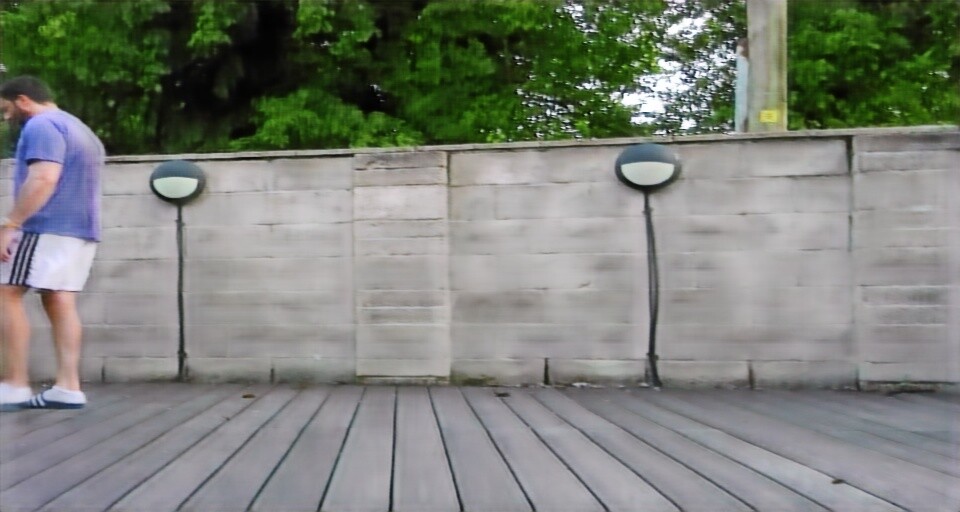} \\

StableLLVE
& \includegraphics[width=\linewidth]{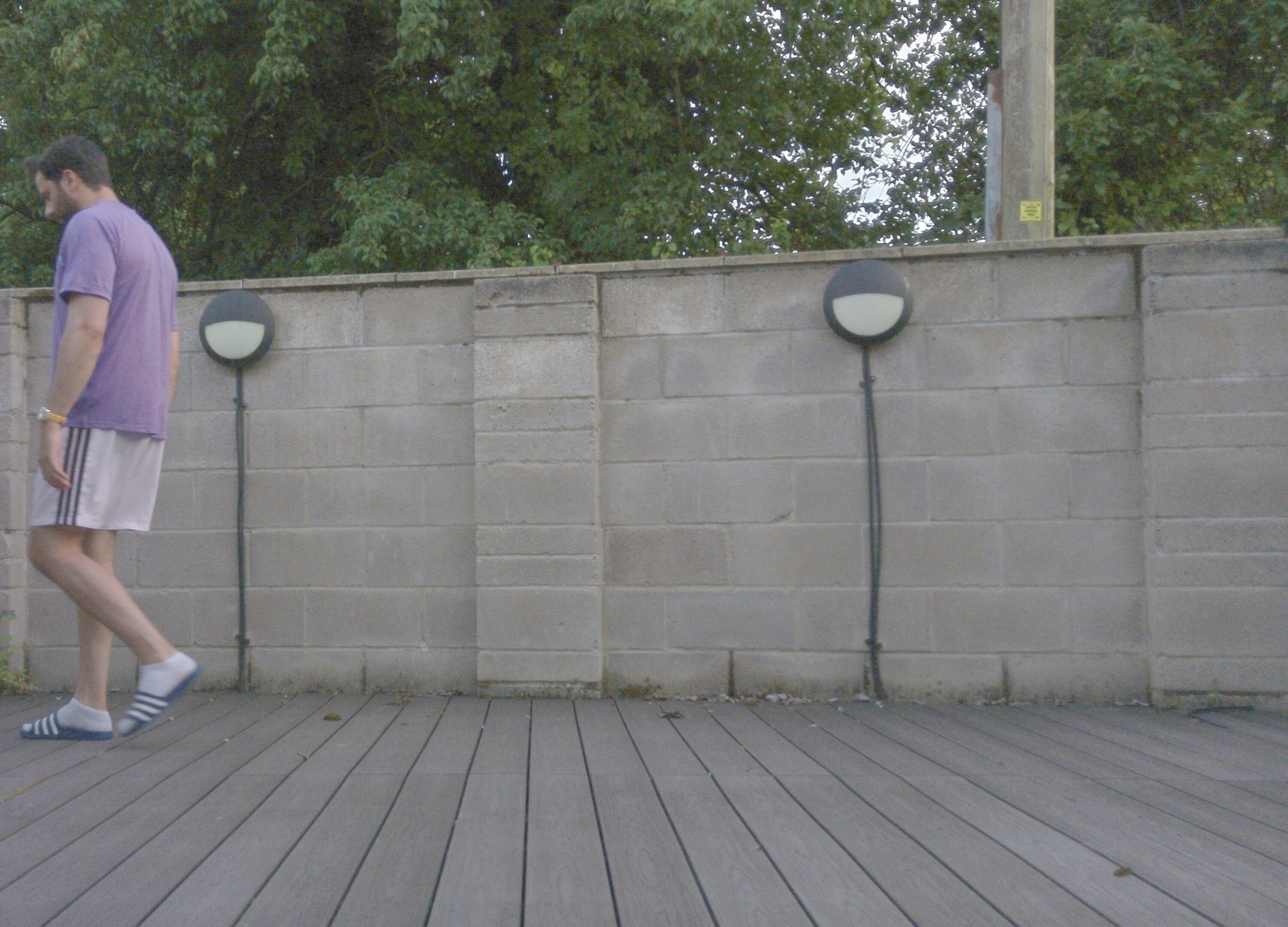}
& \includegraphics[width=\linewidth]{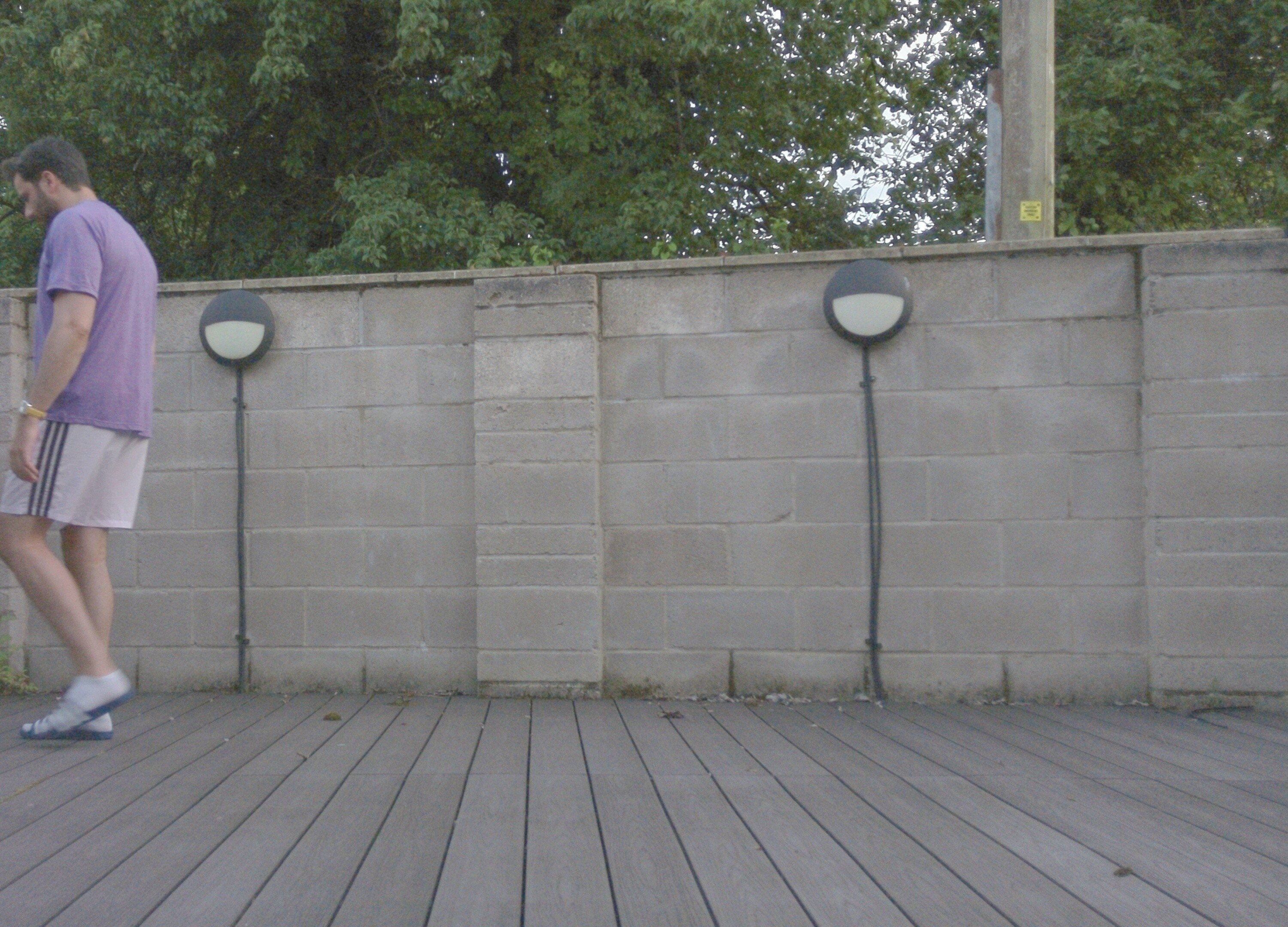}
& \includegraphics[width=\linewidth]{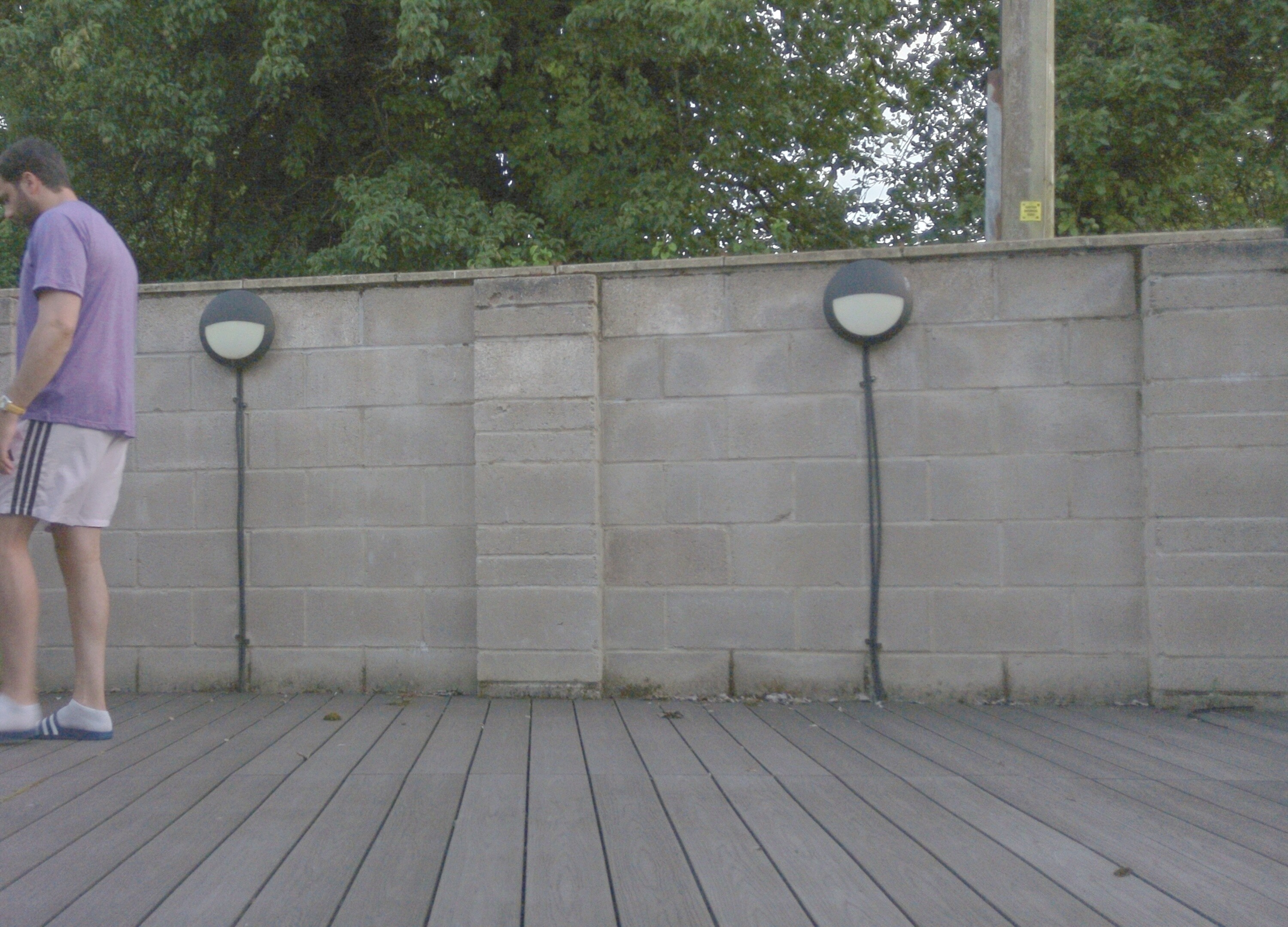} \\

\makecell{SGLLIE/\\SGZ}
& \includegraphics[width=\linewidth]{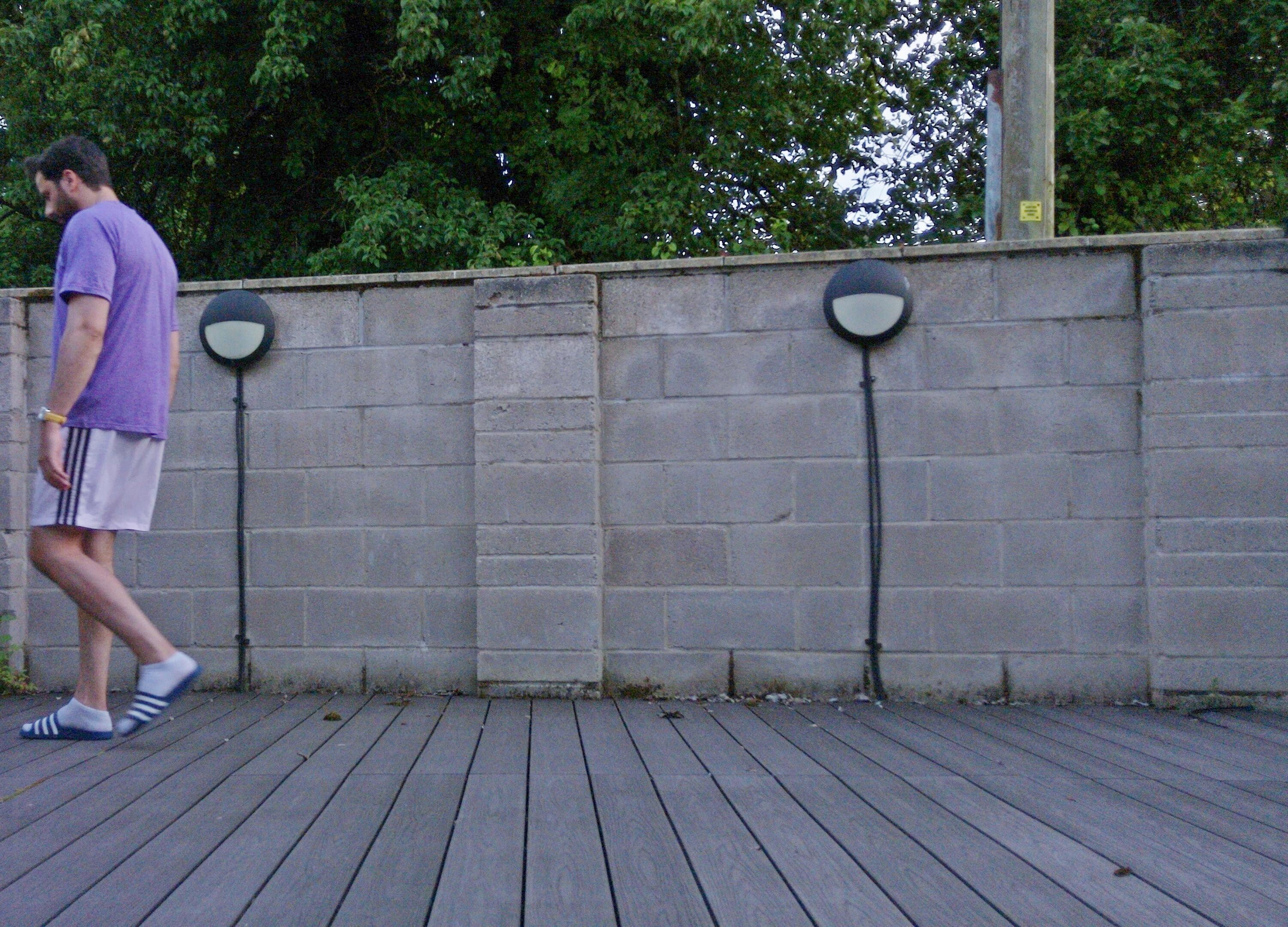}
& \includegraphics[width=\linewidth]{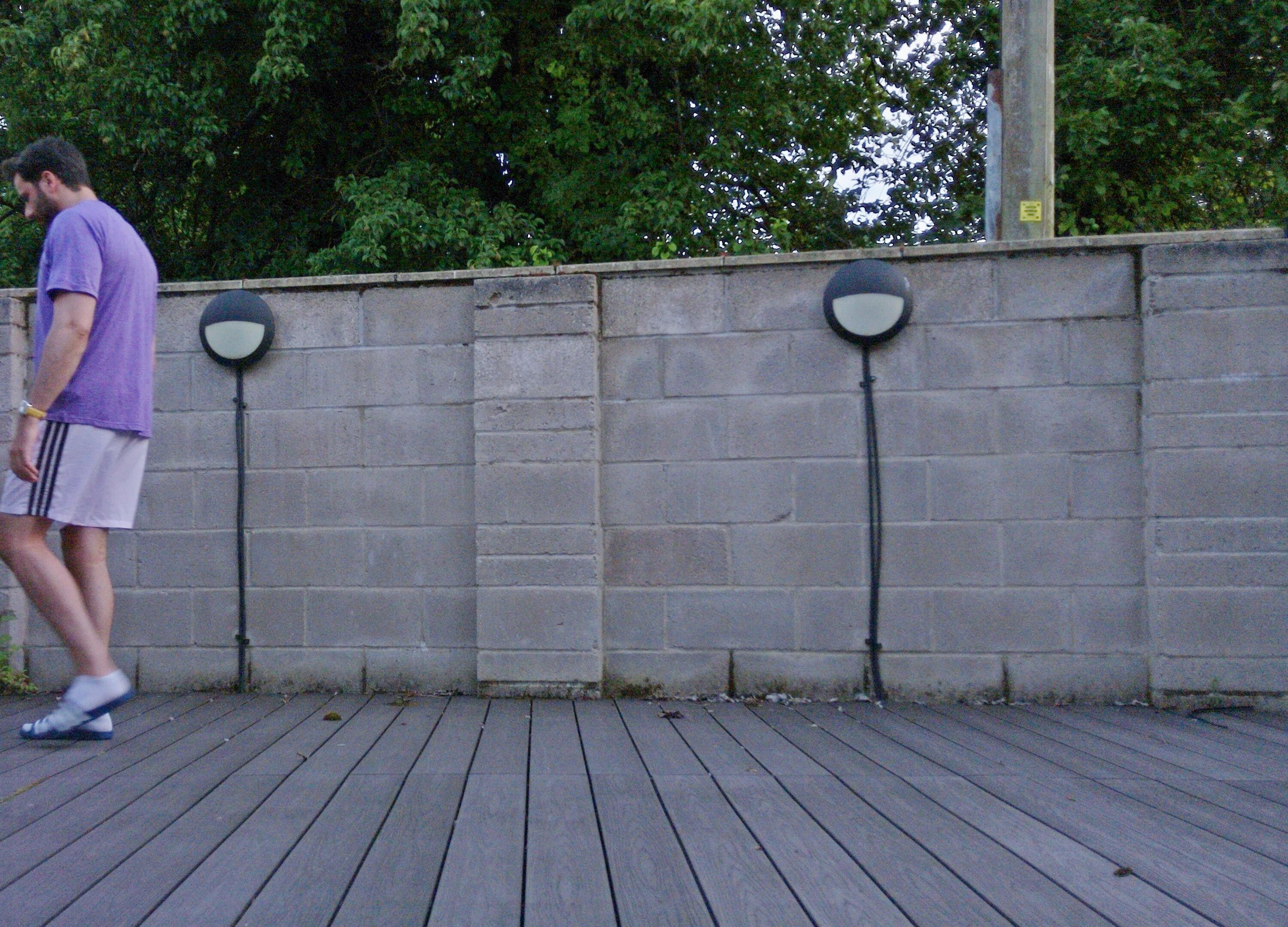}
& \includegraphics[width=\linewidth]{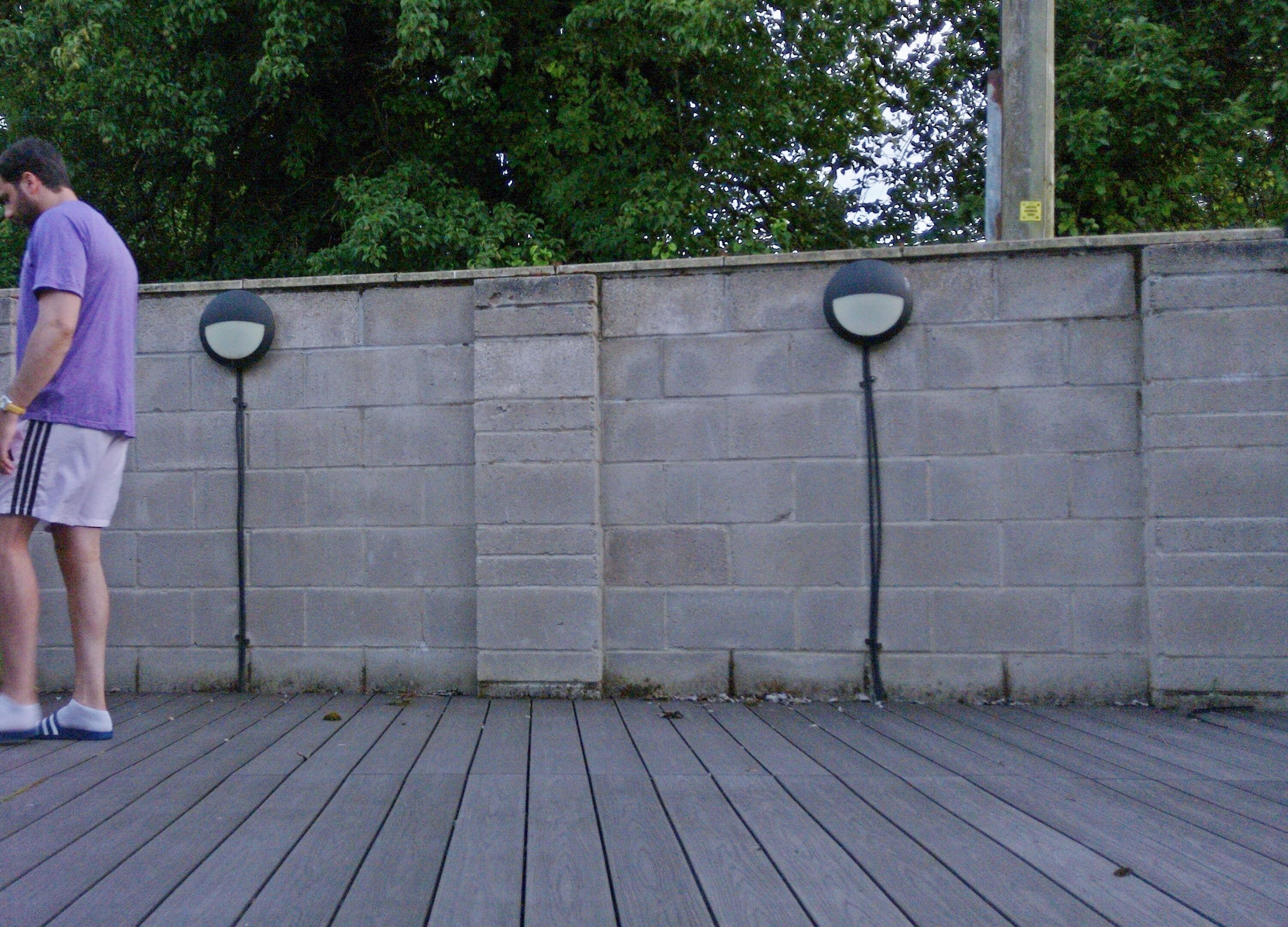} \\

Reference
& \includegraphics[width=\linewidth]{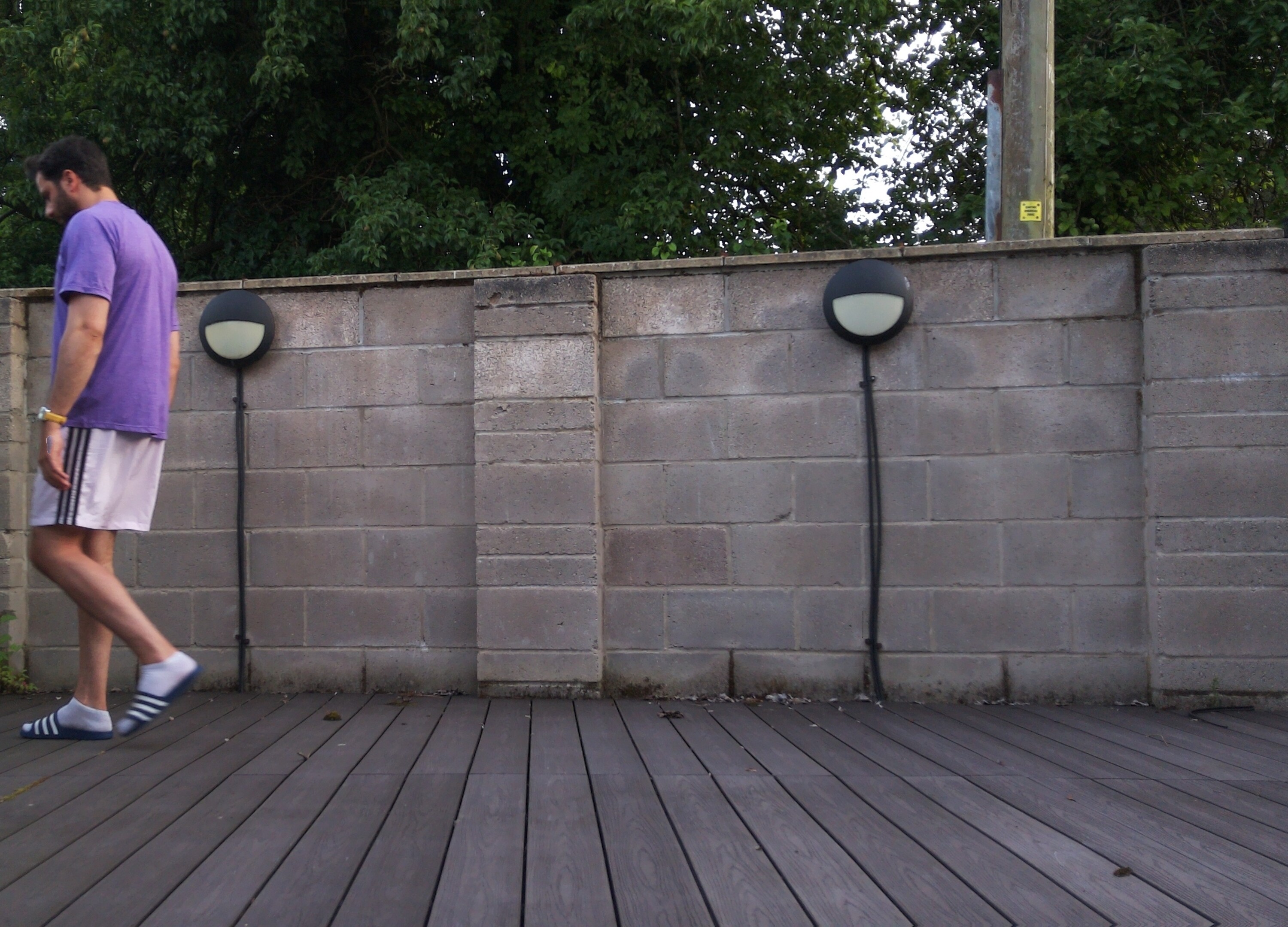}
& \includegraphics[width=\linewidth]{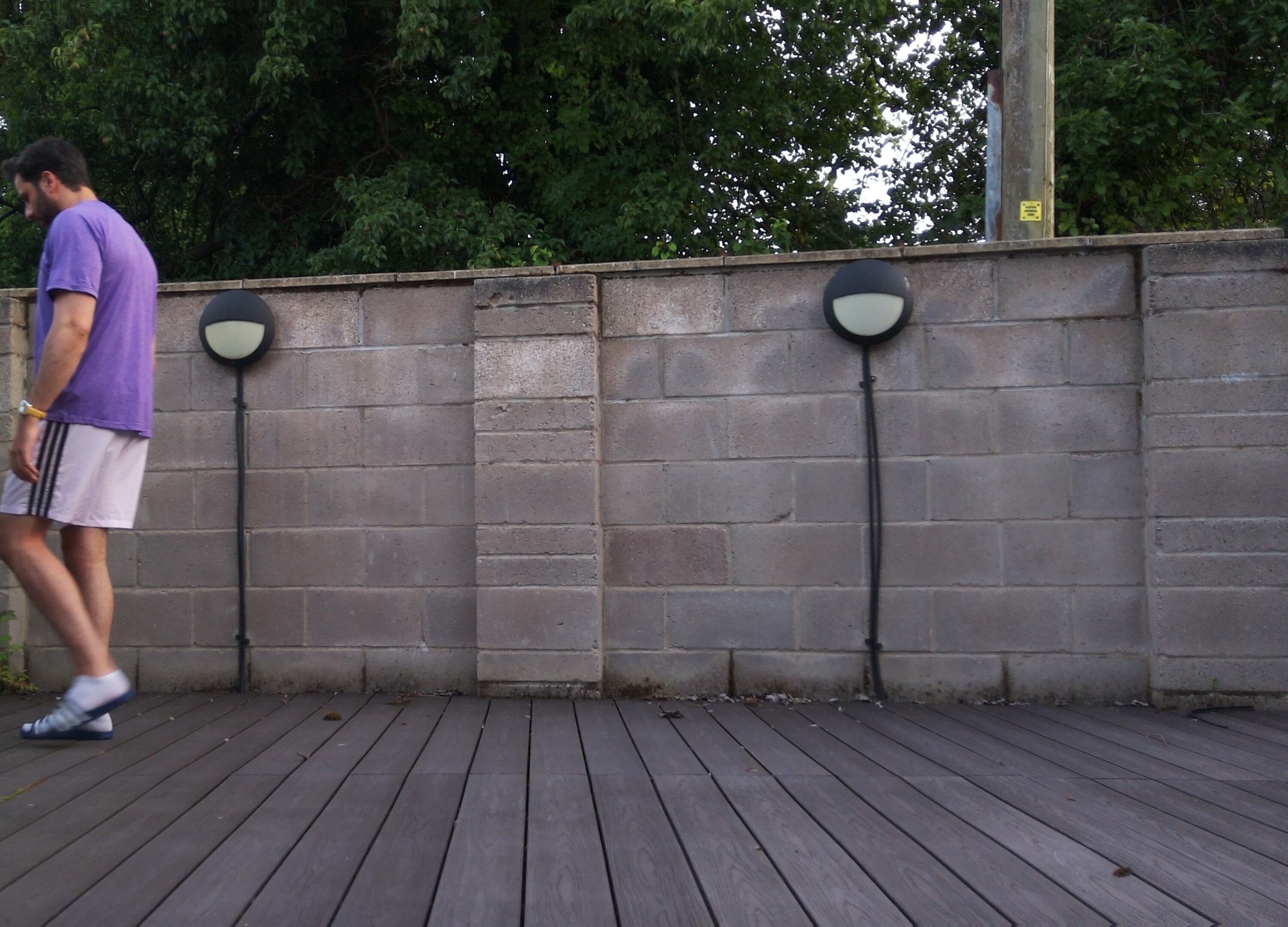}
& \includegraphics[width=\linewidth]{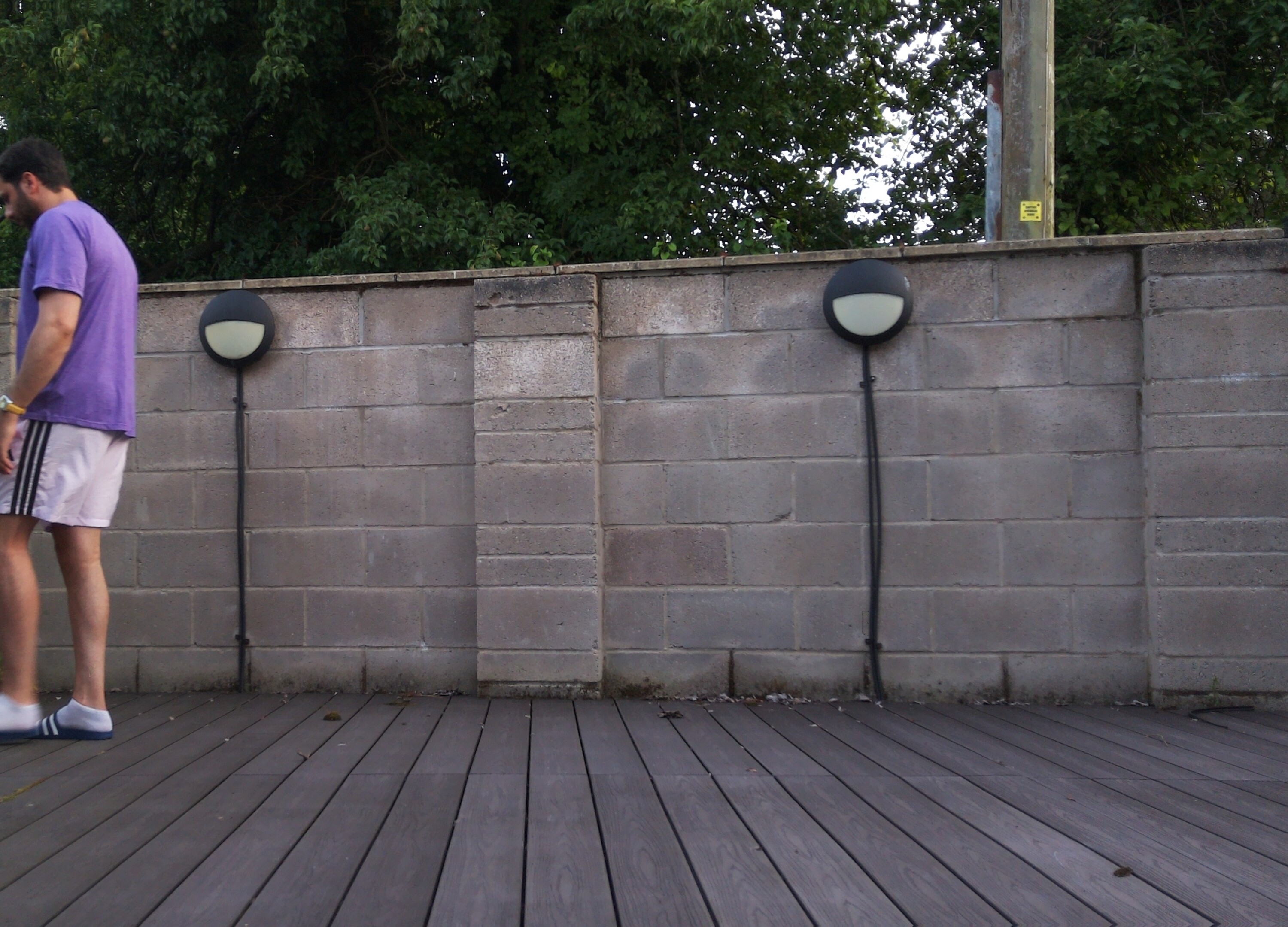} \\

\end{tabular}
}%

\caption{Qualitative comparison on three consecutive frames from a representative outdoor PiLLVD sequence. The comparison includes the low-light input, outputs of the benchmarked enhancement methods, and the aligned normal-light reference.}
\label{fig:qual_outdoor}
\end{figure*}

Figure~\ref{fig:qual_outdoor} presents the outdoor comparison, which is more challenging due to the moving person, larger scene depth, textured background, vegetation, and non-uniform illumination. The input frames are dark, and details of the person, wall, floor, and background trees are weakly visible. Compared with the aligned normal-light reference, the evaluated methods show different trade-offs between brightness recovery, colour fidelity, detail preservation, and motion-related artefacts. SDSDNet substantially increases brightness, but introduces visible ghosting and colour smearing around the moving person, together with over-bright artefacts in the lower-left region. DP3DF produces a more conservative enhancement and preserves the overall scene layout, but the output remains under-enhanced compared with the reference, particularly on the wall and floor. LLVE-STCD improves the visibility of the wall, floor, vegetation, and moving person; however, some regions appear over-bright relative to the reference, and the colour tone remains slightly shifted. StableLLVE generates a bright and smooth result, but the output appears hazy and lower in contrast, with weakened wall and floor texture. SGLLIE/SGZ also improves visibility and preserves the global scene structure, but the output is brighter than the reference in several regions and loses some local texture on the wall and floor. The outdoor example highlights the difficulty of enhancing low-light videos containing natural motion. Compared with the indoor case, the outdoor sequence contains stronger spatial variation, non-uniform illumination, background vegetation, and more complex human motion. The moving person is particularly informative: some methods increase visibility but introduce ghosting, colour distortion, or over-enhancement around the body, while others preserve a more stable appearance but under-enhance the scene. These visual observations are consistent with the quantitative results, where the outdoor sequences show mixed winners across PSNR, SSIM, LPIPS, and mTWE.

Overall, the qualitative results confirm that no single pre-trained method consistently matches the aligned normal-light reference across both indoor and outdoor scenes. Some models are effective at increasing brightness, but may introduce colour shifts, over-enhancement, or unnatural contrast. Others preserve temporal stability or structural information better, but may produce under-enhanced or overly smooth outputs. These results support the quantitative findings and demonstrate that PiLLVD provides a challenging benchmark for evaluating not only frame-level restoration quality, but also perceptual naturalness and temporal consistency in dynamic low-light video enhancement.

\section{Discussion \& Limitations}
\label{Dis}

The proposed PiLLVD dataset is designed to address an important gap in low-light video enhancement research: the lack of paired real-world video sequences containing dynamic scene content. Although PiLLVD is compact in scale, its value lies in the diversity of its capture conditions, scene content, and motion patterns rather than in the total number of frames alone. In contrast to datasets collected under highly controlled or repeated motion settings, PiLLVD contains varied indoor and outdoor scenes with different objects, human activities, backgrounds, and illumination levels. This diversity makes it useful for evaluating whether enhancement methods can generalise beyond static scenes, synthetic degradation, or simple predefined motion patterns, and whether they remain reliable when applied to more natural dynamic scenarios. This is particularly important for low-light video enhancement because realistic movement affects both individual frame quality and consistency between neighbouring frames. A dataset with simple or highly controlled motion may not fully test how enhancement methods behave under natural scene changes. By including different motion sources, scene structures, and illumination conditions, PiLLVD supports the analysis of motion-related artefacts, temporal instability, and robustness under realistic low-light conditions.

The benchmark results demonstrate that existing pre-trained enhancement models do not generalise uniformly to PiLLVD. No single method consistently achieves the best performance across fidelity, perceptual quality, and temporal stability metrics. Some methods improve brightness effectively but introduce colour shifts, noise amplification, over-enhancement, or unnatural contrast. Others preserve structural details or temporal consistency better, but may under-enhance the scene or produce overly smooth results with weakened local texture. These findings suggest that dynamic paired low-light videos introduce challenges that are not fully represented in existing training and evaluation datasets. Therefore, PiLLVD provides a practical dataset for analysing the limitations of current low-light video enhancement models and supporting the development of future methods under realistic dynamic conditions.

A key challenge in constructing PiLLVD is the alignment of paired low-light and normal-light frames. Although the proposed acquisition system provides synchronised capture, the post-capture pipeline is still required to improve spatial correspondence between the paired views before they can be used as reliable references. This pipeline substantially reduces spatial misalignment, in some cases to nearly negligible levels, although perfect pixel-level alignment remains difficult in challenging frames. This is particularly challenging when the low-light input is extremely dark, since enhancing such frames for optical-flow estimation can amplify sensor noise and reduce the reliability of correspondence estimation. Very fast motion also introduces physical motion blur, making it difficult to establish accurate matches between the paired low-light and normal-light frames. These factors can lead to alignment failures or residual local misalignment, especially around moving objects, blurred regions, and low-texture areas.

Another limitation is that the current release contains a finite number of sequences and a larger proportion of indoor scenes than outdoor scenes. While the dataset includes varied motion patterns, objects, backgrounds, and illumination conditions, future versions will expand the number of sequences, with particular emphasis on outdoor scenarios, faster motion, and more challenging low-light conditions. The current benchmark is also limited to zero-shot evaluation of selected publicly available RGB-compatible models. This setting is intentional, as the aim is to assess how existing pre-trained methods generalise to PiLLVD without adaptation to the proposed dataset. In practice, several benchmarked methods require substantial computational resources during inference, and some process frames at reduced spatial resolutions to make enhancement feasible. This highlights the continuing need for efficient low-light video enhancement methods that can operate under real-time or resource-constrained conditions. Training or fine-tuning enhancement models directly on PiLLVD is therefore an important direction for future work, allowing the dataset to support not only generalisation evaluation but also the development of efficient models that better exploit paired dynamic low-light video data.

Overall, PiLLVD demonstrates that realistic dynamic low-light video enhancement remains an open challenge. The dataset provides paired low-light and aligned normal-light reference sequences captured with an accessible acquisition system, while the benchmark highlights the limitations of current methods under unseen dynamic conditions. By combining paired acquisition, post-capture alignment, diverse dynamic content, and zero-shot evaluation, PiLLVD offers a useful resource for both analysing existing enhancement models and encouraging future research on robust low-light video enhancement. 

\section{Conclusion}
\label{con}

This paper introduced PiLLVD, a real-world paired low-light video dataset designed for dynamic scene enhancement. The dataset was collected using a low-cost Raspberry Pi-based dual-camera acquisition system with a beam splitter, enabling synchronised capture of paired low-light and normal-light video sequences using accessible off-the-shelf components. Unlike many existing low-light datasets that are image-based, static, synthetic, unavailable, or constrained by simple motion patterns, PiLLVD provides paired dynamic indoor and outdoor scenes with diverse human and object motion, varied backgrounds, and different illumination conditions.

In addition to the dataset, we presented a post-capture alignment pipeline to improve spatial correspondence between paired low-light and normal-light frames and evaluated the alignment quality using feature-based geometric metrics. We also benchmarked representative pre-trained low-light video enhancement models on PiLLVD under a zero-shot setting. The quantitative and qualitative results show that existing methods do not generalise uniformly to dynamic real-world low-light videos, with different models exhibiting trade-offs between fidelity, perceptual quality, and temporal stability.

Overall, PiLLVD provides a practical and challenging benchmark for analysing the limitations of current low-light video enhancement methods and supporting the development of more robust models. Future work will extend the dataset with additional sequences, particularly outdoor scenes and more challenging low-light conditions, and investigate efficient enhancement models trained directly on PiLLVD for real-time or resource-constrained deployment.

\bibliographystyle{IEEEtran}
\bibliography{ref}

\end{document}